\documentclass[conference]{IEEEtran}
\IEEEoverridecommandlockouts
\pdfoutput=1
\usepackage{ragged2e}
\usepackage{makecell,booktabs,tabularx}
\usepackage{bigstrut,multirow,rotating}
\usepackage{threeparttable}
\usepackage{lineno}
\usepackage{enumitem}
\usepackage{cite}
\usepackage{amsmath,amsthm,amssymb,amsfonts,bm}
\usepackage{algorithm,algorithmicx}
\usepackage{algpseudocode}
\usepackage{graphicx,subfigure,svg}
\usepackage{textcomp}
\usepackage{xcolor}
\def\BibTeX{{\rm B\kern-.05em{\sc i\kern-.025em b}\kern-.08em
    T\kern-.1667em\lower.7ex\hbox{E}\kern-.125emX}}

\usepackage{threeparttable}

\newtheorem{definition}{Definition}[section] % 定义Definition的环境
\newtheorem{assumption}{Assumption}[section] % 定义assumption的环境
\begin{document}

% \pagewiselinenumbers% 按页重新编号 
% \switchlinenumbers	% 双栏

\title{Stable Heterogeneous Treatment Effect Estimation across Out-of-Distribution Populations}

% 作者居中
\makeatletter
\newcommand{\linebreakand}{%
  \end{@IEEEauthorhalign}
  \hfill\mbox{}\par
  \mbox{}\hfill\begin{@IEEEauthorhalign}
}
\makeatother

\author{\IEEEauthorblockN{Yuling Zhang\IEEEauthorrefmark{1},
Anpeng Wu\IEEEauthorrefmark{3}\IEEEauthorrefmark{4},
Kun Kuang\IEEEauthorrefmark{3}, 
Liang Du\IEEEauthorrefmark{5},
Zixun Sun\IEEEauthorrefmark{5}, and
Zhi Wang\IEEEauthorrefmark{1}\IEEEauthorrefmark{2}}
\IEEEauthorblockA{\IEEEauthorrefmark{1}Tsinghua Shenzhen International Graduate School, Tsinghua University}
\IEEEauthorblockA{\IEEEauthorrefmark{2}Tsinghua-Berkeley Shenzhen Institute, Tsinghua University}
\IEEEauthorblockA{\IEEEauthorrefmark{3}College of Computer Science and Technology, Zhejiang University}
\IEEEauthorblockA{\IEEEauthorrefmark{4}Machine Learning Department, Mohamed bin Zayed University of Artificial Intelligence}
\IEEEauthorblockA{\IEEEauthorrefmark{5}Interactive Entertainment Group, Tencent}
\IEEEauthorblockA{Email: zhangyl21@mails.tsinghua.edu.cn, \{anpwu,kunkuang\}@zju.edu.cn,}
\IEEEauthorblockA{\{lucasdu,zixunsun\}@tencent.com, wangzhi@sz.tsinghua.edu.cn}
% <-this % stops an unwanted space
\thanks{Kun Kuang and Zhi Wang are the corresponding authors.}}

% \textit{Tsinghua-Berkeley Shenzhen Institute} \\
% \textit{College of Computer Science \& Technology} \\
% \textit{Interactive Entertainment Group} \\

% \author{\IEEEauthorblockN{Yuling Zhang}
% \IEEEauthorblockA{\textit{Tsinghua University} \\
% zhangyl21@mails.tsinghua.edu.cn}
% \and
% \IEEEauthorblockN{Anpeng Wu}
% \IEEEauthorblockA{\textit{Zhejiang University \& MBZUAI} \\
% anpwu@zju.edu.cn}
% \and
% \IEEEauthorblockN{Kun Kuang}
% \IEEEauthorblockA{\textit{Zhejiang University} \\
% kunkuang@zju.edu.cn}
% \linebreakand
% \IEEEauthorblockN{Liang Du}
% \IEEEauthorblockA{\textit{Tencent} \\
% lucasdu@tencent.com}
% \and
% \IEEEauthorblockN{Zixun Sun}
% \IEEEauthorblockA{\textit{Tencent} \\
% zixunsun@tencent.com}
% \and
% \IEEEauthorblockN{Zhi Wang}
% \IEEEauthorblockA{\textit{Tsinghua University} \\
% wangzhi@sz.tsinghua.edu.cn}
% \thanks{Kun Kuang and Zhi Wang are the corresponding authors.}
% }

\maketitle

\begin{abstract}
Heterogeneous treatment effect (HTE) estimation is vital for understanding the change of treatment effect across individuals or subgroups. Most existing HTE estimation methods focus on addressing selection bias induced by imbalanced distributions of confounders between treated and control units, but ignore distribution shifts across populations. Thereby, their applicability has been limited to the in-distribution (ID) population, which shares a similar distribution with the training dataset. In real-world applications, where population distributions are subject to continuous changes, there is an urgent need for stable HTE estimation across out-of-distribution (OOD) populations, which, however, remains an open problem. As pioneers in resolving this problem, we propose a novel Stable Balanced Representation Learning with Hierarchical-Attention Paradigm (SBRL-HAP) framework, which consists of 1) Balancing Regularizer for eliminating selection bias, 2) Independence Regularizer for addressing the distribution shift issue, 3) Hierarchical-Attention Paradigm for coordination between balance and independence. In this way, SBRL-HAP regresses counterfactual outcomes using ID data, while ensuring the resulting HTE estimation can be successfully generalized to out-of-distribution scenarios, thereby enhancing the model's applicability in real-world settings. Extensive experiments conducted on synthetic and real-world datasets demonstrate the effectiveness of our SBRL-HAP in achieving stable HTE estimation across OOD populations, with an average $10\%$ reduction in the error metric PEHE and $11\%$ decrease in the ATE bias, compared to the SOTA methods.

\end{abstract}

\begin{IEEEkeywords}
Heterogeneous Treatment Effect; Out-of-Distribution; Balanced Representation Learning; Hierarchical-Attention Optimization 
\end{IEEEkeywords}

\section{Introduction}

\begin{figure}[t]

\centering
\includegraphics[width=0.95\linewidth]{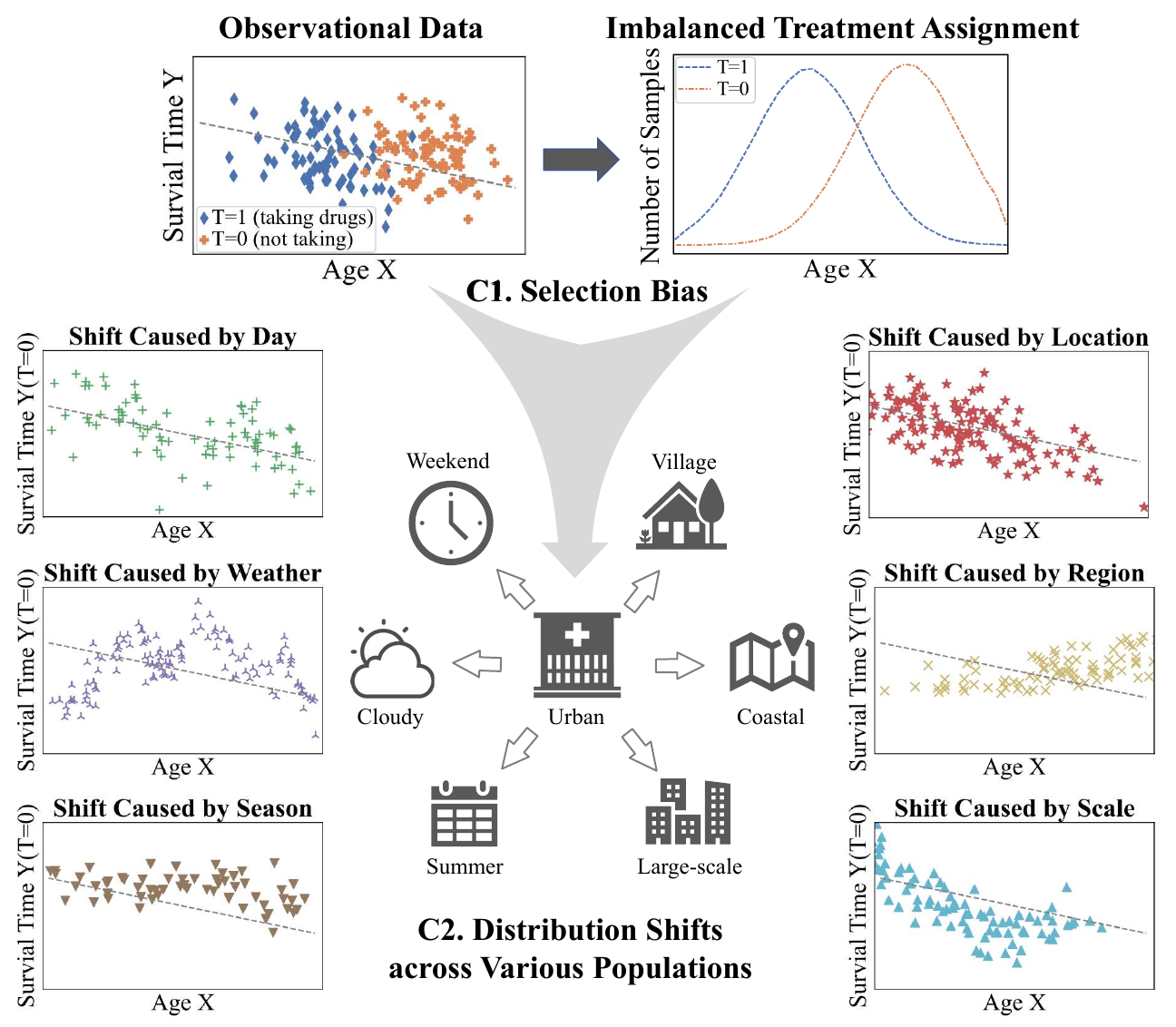}
\caption{Two main challenges in stable HTE estimation across OOD populations: \textbf{(C1)} selection bias from imbalanced treatment assignment, and \textbf{(C2)} distribution shift across various populations. The former is manifested as imbalanced distributions of covariates (e.g., age) between treated (i.e., T=1) and control (i.e., T=0) units in a specific population. The latter occurs frequently in real-world applications, resulting in out-of-distribution populations that have distinct covariate distributions from the training dataset. This work is among the first to synergistically resolve both selection bias and distribution shift.}
\label{fig:intro}
\end{figure}

Estimating Heterogeneous Treatment Effects (HTE) from observational data has gained increasing importance across various fields~\cite{chuContinualCausalInference2023}, including medicine, economics, and marketing~\cite{aiLBCFLargeScaleBudgetConstrained2022,tanUncoveringCausalEffects2022,mengCausalAnalysisUnsatisfying2019,wagerEstimationInferenceHeterogeneous2018}. This can provide practitioners valuable insights into understanding how treatment effects vary among different subpopulations, ultimately achieving personalized health-care and explainable decision-making. However, reliable and robust estimation of HTE still faces significant challenges.
One primary challenge in observational data is non-random treatment assignment, which can lead to imbalanced covariate distributions between treated and control units (Top panel in Fig.~\ref{fig:intro}). Taking healthcare as an example, in the study of the effect of treatment on outcomes, physicians may assign different treatment recommendations (e.g., taking the drug or not) based on the patient's individual circumstances (e.g., age). Typically, physicians recommend young individuals to take the medication more often while advising older individuals not to take it. Such imbalanced treatment assignment can result in selection bias~\cite{pearlCausality2009}, manifested as differences in age distribution between the treated group and the control group. 
As selection bias has been taken seriously by academia and industry~\cite{wangSequentialRecommendationUser2023,zhuDCMTDirectEntireSpace2023,youngmannExplainingConfoundingBias2023,shenCausalWhatIfHowTo2023}, various methods such as propensity score matching, doubly robust, stratification, inverse probability of treatment weighting (IPTW)~\cite{rosenbaumCentralRolePropensity1983,rosenbaumModelBasedDirectAdjustment1987,liMatchingDimensionalityReduction2016,yaoSurveyCausalInference2021}, and representation learning methods~\cite{shalitEstimatingIndividualTreatment2017,hassanpourCounterFactualRegressionImportance2019,wuLearningDecomposedRepresentations2022,hassanpourLearningDisentangledRepresentations2020,schwabLearningCounterfactualRepresentations2020a,yaoSCISubspaceLearning2021,yaoACEAdaptivelySimilarityPreserved2019,yaoRepresentationLearningTreatment2018} have been developed to reduce selection bias and estimate treatment effects more accurately. 

However, one limitation is that these methods have only been tested and validated on data that is similar to the training data, known as in-distribution (ID) data. In real-world applications, where data or population distributions, specifically the covariate distributions, are subject to continuous changes~\cite{zhangDiversePreferenceAugmentation2022,caoCrossDomainRecommendationColdStart2022,zhouActiveGradualDomain2022,zhouOnlineContinualAdaptation2022,fangGIFDGenerativeGradient2023}, there is a concern regarding the performance of these methods when applied to populations with different covariate distributions compared to the training dataset~\cite{wuIterativeRefinementMultiSource2023,yanTransferableFeatureSelection2023,chenUnsupervisedIntraDomainAdaptation2023,chenBAGNNLearningBiasAware2022,yuanLabelEfficientDomainGeneralization2022}.
This issue, referred to as distribution shift~\cite{sugiyamaCovariateShiftAdaptation2007,shimodairaImprovingPredictiveInference2000}, has posed another significant challenge to achieving stable HTE estimation for out-of-distribution (OOD) populations.
As shown in Fig.~\ref{fig:intro}, the distribution of patients' circumstances may change over time, seasons, holidays, urban and rural areas, etc., resulting in the emergence of various populations.
These populations may have different data distributions and characteristics compared to the training data, and they may even include individuals that were not present in the training data. 
Due to induction bias, the causal relations learnt from training data (e.g., data collected during weekdays) are typically not applicable to testing data (e.g., data collected during weekends). If we directly use the above causal models trained on one specific dataset, it may lead to unstable and unreliable HTE estimation for other populations.
% \textcolor{blue}{Traditional methods may train and test causal models on similar populations, without taking distribution changes into account. If we directly use these trained causal models to estimate HTE for the populations with varying data distributions and characteristics compared to the training data, potentially including individuals not present in the training data, it may lead to unreliable HTE estimation for these OOD populations.}
Such unreliability of HTE estimation can lead to inappropriate treatment choices, posing a huge threat to patients' health and even resulting in catastrophic medical events. Therefore, there is an urgent demand to develop stable HTE estimation methods that can effectively generalize to unseen samples or different populations.

In this paper, we first study the problem of stable HTE estimation across OOD populations, and systemically review the two main challenges (Fig.~\ref{fig:intro}): \textbf{(C1)} selection bias from imbalanced treatment assignment, and \textbf{(C2)} distribution shift across various populations. 
Selection bias in observational data can lead to unreliable and biased HTE estimation. Although many previous causal methods have been proposed to eliminate selection bias, they still suffer from the distribution shift issue, resulting in a higher error and unstable estimates of HTE on out-of-distribution populations.
% In this paper, we first study the problem of stable HTE estimation across various populations, and systemically review the two main challenges (Figure~\ref{fig:intro}): \textbf{(C1)} selection bias from imbalanced treatment assignment, and \textbf{(C2)} distribution shift across various populations.

To address the selection bias, 
\emph{Balanced Representation Learning} (BRL) has been developed to map the original covariates to a representation space and narrow the representation discrepancies across different treatment arms~\cite{shalitEstimatingIndividualTreatment2017,hassanpourCounterFactualRegressionImportance2019,wuLearningDecomposedRepresentations2022}. This approach enables accurate HTE estimation within the in-distribution data.
Nevertheless, in the presence of distribution shifts across various populations, the problem of stable HTE estimation across OOD populations remains relatively unexplored. Among the many OOD generalization algorithms, \emph{Stable Learning} (SL) stands out as a promising approach~\cite{cuiStableLearningEstablishes2022,wangLearningRobustRepresentations2019a,muandetDomainGeneralizationInvariant2013a} based on the following observation. 
For general machine learning models, model crashes under distribution shifts are mainly caused by the unstable correlation between irrelevant features and the target outcome. This kind of unstable correlation fundamentally stems from the statistical dependence between relevant and irrelevant features~\cite{fanGeneralizingGraphNeural2021a,zhangDeepStableLearning2021,lakeBuildingMachinesThat2017}. Therefore, to address distribution shift and maintain performance across OOD data, SL methods propose to decorrelate all features by sample reweighting, facilitating models to recognize stable and invariant relationships between features and outcomes.

Building upon these methods, we propose a novel framework called Stable Balanced Representation Learning with Hierarchical-Attention Paradigm (SBRL-HAP), which comprises three core components: (a) Balancing Regularizer (BR) to eliminate selection bias and obtain balanced representations; (b) Independence Regularizer (IR) to reweight samples and enforce independencies between features, addressing the distribution shift issue; (c) Hierarchical-Attention Paradigm (HAP) to assign distinct priorities to each neural network layers for comprehensive feature decorrelation throughout the learning process. Notably, in the training process of BR and IR, the learning of balanced representation and independence-driven weights can be interdependent. For instance, when representations change, the learning of weights would also adapt accordingly. In such cases, optimizing one objective may come at the expense of the other. Consequently, we design a Hierarchical-Attention Paradigm to synergistically facilitate the learning of balanced representations and independence-driven weights, thereby alleviating conflicts.
To differentiate, we refer to the model without the Hierarchical-Attention Paradigm as SBRL.

The primary contributions of this paper are threefold:
\begin{itemize}
\item In this paper, we first investigate the problem of stable heterogeneous treatment effect estimation across out-of-distribution populations and pioneer the integration of representation balancing and stable training techniques. 
\item We propose a novel SBRL-HAP framework in which the Hierarchical-Attention Paradigm eliminates selection bias and addresses distribution shifts through comprehensive decorrelation in a hierarchical manner. This flexible framework enables the extension of any existing representation balancing method to various OOD populations. 
% \item We propose a novel SBRL-HAP framework which leverages the Hierarchical-Attention Paradigm to eliminate selection bias and address distribution shift in a comprehensive and cohesive manner. This flexible paradigm enables the extension of any existing representation balancing method to various OOD populations. 
\item Extensive experiments conducted on synthetic and real-world data demonstrate the effectiveness of our SBRL-HAP in achieving stable HTE estimation across OOD populations, compared to the SOTA methods. On the OOD datasets, our SBRL-HAP reduces the error metric PEHE by $10\%$ on average compared with the best baseline, and reduces the ATE bias by up to $14\%$.
\end{itemize}

\section{Related Work}

\textbf{Representation Balancing to Mitigate Selection Bias}. 
Many prior works have concentrated on addressing the challenges of estimating heterogeneous treatment effects from observational data while mitigating selection bias, with a promising method being balanced representation learning~\cite{yaoSurveyCausalInference2021,wuLearningInstrumentalVariable2023}. 
This method minimizes the distribution distance between treated and control groups, effectively balancing confounders and producing similar distributions in the representation space, ultimately improving prediction accuracy for heterogeneous treatment effects.
Specifically, representation balancing methods can be broadly categorized into five groups: 
1) Fundamental methods, such as CFR~\cite{shalitEstimatingIndividualTreatment2017,johanssonLearningRepresentationsCounterfactual2016}, which learn balanced representation by directly minimizing distribution distance between the treated and control groups; 
2) Reweighting methods, such as RCFR~\cite{johanssonLearningWeightedRepresentations2018} and CFR-ISW~\cite{hassanpourCounterFactualRegressionImportance2019}, which incorporate information from treatments and use importance sampling techniques to further mitigate the negative impact of selection bias; 
3) Similarity-based methods, such as SITE~\cite{yaoRepresentationLearningTreatment2018} and ACE~\cite{yaoACEAdaptivelySimilarityPreserved2019}, which focus on learning balanced representations while preserving similarity information among data points; 
4) Subgroup methods, such as HNN~\cite{changInformativeSubspaceLearning2017} and SCI~\cite{yaoSCISubspaceLearning2021}, which enhance the model's predictive ability by identifying and partitioning sub-spaces within the representation; 
and 5) Decomposition methods, such as DR-CFR~\cite{hassanpourLearningDisentangledRepresentations2020} and DeR-CFR~\cite{wuLearningDecomposedRepresentations2022}, which separate confounders from pre-treatment variables to achieve precise balancing of covariates. These methods have proven successful in estimating treatment effects without taking distribution shifts into account, but they may be prone to performance degradation in OOD scenarios.

\textbf{Stable Learning to Eliminate Distribution Shifts}. 
Distribution shifts across distinct populations in HTE estimation are not as well explored, and stable learning is a promising approach to address the distribution shift issue~\cite{cuiStableLearningEstablishes2022}.
Taking inspiration from variable balancing strategies in causal inference~\cite{kuangEstimatingTreatmentEffect2017,atheyApproximateResidualBalancing2018,hainmuellerEntropyBalancingCausal2012}, stable learning eliminates dependence among covariates via sample reweighting to manifest causation, thus utilizing the stability of causation to achieve generalization.
% Another closely related line of research is stable learning, which aims to tackle the challenges of cross-temporal and cross-platform distribution shifts \cite{cuiStableLearningEstablishes2022}. 
Recently, several studies, including~\cite{shenCausallyRegularizedLearning2018, kuangStablePredictionModel2020, shenStableLearningSample2020, zhangDeepStableLearning2021}, have aimed to tackle the discrepancy between the training and testing distribution stemming from datasets collected at different time periods or platforms.
These approaches have the potential to handle distribution shifts in HTE estimation. 
Among them, CRLR~\cite{shenCausallyRegularizedLearning2018} addresses distribution shifts by simultaneously optimizing global confounder balancing and weighted logistic regression to estimate the causal effect of each variable on the outcome. However, CRLR requires that all the features and labels be binary, which is impractical in real-world applications.
To overcome this limitation, DWR~\cite{kuangStablePredictionModel2020} proposes to utilize the statistical independence condition to force that variables are independent of each other, thereby relaxing the binary restriction.
Furthermore, SRDO~\cite{shenStableLearningSample2020} constructs an uncorrelated design matrix from original covariates to alleviate the issue of co-linearity among variables.
On the other hand, StableNet~\cite{zhangDeepStableLearning2021} goes beyond the linear case and addresses both linear and non-linear dependencies between variables using Random Fourier Features and the Hilbert-Schmidt Independence Criterion.
Overall, stable learning techniques aim to realize model generalization across any distribution by excavating stable relationships through feature decorrelation.

Although Representation Balancing \cite{shalitEstimatingIndividualTreatment2017} and Stable Learning \cite{zhangDeepStableLearning2021} can address Selection Bias from imbalanced treatment assignment and distribution shift across data respectively, their optimization objectives are not orthogonal. The learning of weights and representations can interfere with each other, which is the reason why few works have discussed Stable Estimation in HTE across data. Considering the increasing importance of stable HTE estimation, this work pioneers to propose a novel framework named SBRL-HAP, in which the Hierarchical-Attention Paradigm coordinates the Balancing Regularizer and Independence Regularizer to extract balanced and stable representations, thus bridging these two topics. 

\section{Problem Setup and Assumptions} \label{sec:prb}

\subsection{Problem Setup}

In this paper, we study the heterogeneous treatment effect estimation across multiple populations. For simplicity, we consider that the population used for training the model is drawn from environment $e \in \mathcal{E}$ and the target population is from environment $e' \in \mathcal{E}$. 
% In this paper, we study the Heterogeneous Treatment Effect Estimation  across multiple environments or populations. For simplicity, let's denote the population used for training the model as $s \in \mathcal{E}$ and the target population as $v \in \mathcal{E}$. 
Taking healthcare as an example, as illustrated in Fig.~\ref{fig:intro}, we gather an observational  $D^e=\{\mathbf{X}^e,T^e,Y^e\}=\{\mathbf{x}_i^e,t_i^e,y_i^{t_i,e}\}_{i=1}^{n}$ from urban hospitals represented by the environment $e$, where $\mathbf{x}^e_i \in \mathcal{X}$ denotes the covariates (e.g., patients circumstances), $t_i^e \in \{0, 1\}$ denotes the received treatment (e.g., take drug or not), and $y_i^{t_i,e} \in \mathcal{Y}$ is the observed outcome corresponding to the treatment $t_i^e$. 
Then, in the target environment $e' \in \mathcal{E}$ different from $e$, such as a remote village, we have a potential population denoted as $D^{e'}=\{\mathbf{X}^{e'}\}$. This dataset only includes the covariates $\mathbf{x}$ of the individuals in the target population, without the corresponding treatment or outcome information. 
Our goal is to learn a causal model from the dataset $D^e$ which enables accurate HTE estimations for the target populations from different environments $e' \in \mathcal{E}$.
% Our goal is to utilize the heterogeneous treatment effect (HTE) estimation obtained from training on the urban hospitals' data to make accurate predictions for individuals in different populations with distinct distributions.
We refer to this problem as Heterogeneous Treatment Effect Estimation across Out-of-Distribution Populations.
% This problem is referred to as Heterogeneous Treatment Effect Estimation across Out-of-Distribution Populations.

Our work focuses on the Heterogeneous Treatment Effect at the individual level, i.e., Individual Treatment Effect (ITE), and Average Treatment Effect (ATE) at the population level.

\begin{definition}[Individual Treatment Effect]
Given any environment $e \in \mathcal{E}$, the Individual Treatment Effect of unit $i$ is:
\begin{equation}
ITE_i^{e}=y_i^{1,e}-y_i^{0,e},
\end{equation}
where $y_i^{1,e}$ and $y_i^{0,e}$ are potential outcomes.
\end{definition}

\begin{definition}[Average Treatment Effect]
Given any environment $e \in \mathcal{E}$, the Average Treatment Effect of $D^{e}$ is:
\begin{equation}
ATE^{e}=\mathbb{E}[Y^{1,e}-Y^{0,e}]=\frac{1}{n}\sum_{i=1}^n(y_i^{1,e}-y_i^{0,e}).
\end{equation}
\end{definition}

\subsection{Assumptions}

Given the training data $D^e=\{\mathbf{X}^e,T^e,Y^e\}$ from environment $e$, our goal is to find a regressor $f(\cdot):\mathcal{X} \times \mathcal{T} \rightarrow \mathcal{Y}$ capable of precisely predicting potential outcomes across different OOD environments $e' \in \mathcal{E}$. 
To eliminate the selection bias in $D^e$, existing causal models rely on standard assumptions~\cite{guidowimbensCausalInferenceStatistics2015}.

\begin{assumption}[Stable Unit Treatment Value]
The distribution of the potential outcome of one unit is assumed to be independent of the treatment assignment of another unit.
\end{assumption}

\begin{assumption}[Unconfoundedness]
The distribution of treatment is independent of the potential outcome when given covariates. Formally, $T \bot (Y^0,Y^1)|\mathbf{X}$.
\end{assumption}

\begin{assumption}[Overlap]
Every unit should have a nonzero probability to receive either treatment status. Formally, $0<p(T=1|\mathbf{X})<1$.
\end{assumption}

Additionally, without any prior knowledge or structural assumptions, it is impossible to figure out the distribution shift problem, since one cannot characterize the rare or unseen latent environments~\cite{liuHeterogeneousRiskMinimization2021}. Thereby, we follow the assumption commonly used in studies of distribution shift~\cite{xuTheoreticalAnalysisIndependencedriven2022,liuHeterogeneousRiskMinimization2021,kuangStablePredictionModel2020}.

\begin{assumption}[Stable Representation]\label{asump:StableFeature}
There exists a stable representation $\Psi^s(\mathbf{X})$ of covariates $\mathbf{X}$ such that for any environment $e \in \mathcal{E}$, $\mathbb{E}[Y,T|\mathbf{X}^{e}]=\mathbb{E}[Y,T|\Psi^s(\mathbf{X}^{e})]$ holds.
\end{assumption}

This assumption implies covariates $\mathbf{X}$ include two parts: relevant features having causal effects on outcome $Y$, known as stable features $\mathbf{X}_S$; And irrelevant features (i.e., unstable features $\mathbf{X}_V$) that have $P^e(Y|\mathbf{X}_V)\neq P^{e'}(Y|\mathbf{X}_V)$ and create instability for prediction.
The existence of $\mathbf{X}_S$ provides the possibility of precisely predicting the outcome $Y$ using $\Psi^s(\mathbf{X})$, which is known as the stable representation with invariant relationships to the outcome $Y$ across different environments $e \in \mathcal{E}$~\cite{liuHeterogeneousRiskMinimization2021,xuTheoreticalAnalysisIndependencedriven2022}.

% This assumption implies the possibility of precisely predicting the outcome $Y$ using $\Psi^s(\mathbf{X})$, which is known as stable representations with invariant relationships to the outcome $Y$ across different environments $e \in \mathcal{E}$~\cite{liuHeterogeneousRiskMinimization2021,xuTheoreticalAnalysisIndependencedriven2022}.

% \vspace{8pt}

% [Suggestion] 对数据层面的问题并没有突出, 或许应该花费更多的篇幅描述真实数据中OOD问题带来的数据方面的考量和问题？然后再引出OOD? 然后我们的方法看作是数据OOD的post-process方法？

\textbf{Challenges}. 
Overall, we formally discuss challenges in stable HTE estimation across OOD populations. \textbf{(C1)} Selection bias refers to the inconsistent distribution of covariates between different treatment arms in a specific environment $e$, i.e., $P^{e}(\mathbf{X}^t)\neq P^{e}(\mathbf{X}^c)$, where $\mathbf{X}^t=\{\mathbf{x}_{i:t_i=1}\}$ and $\mathbf{X}^c=\{\mathbf{x}_{i:t_i=0}\}$. \textbf{(C2)} Distribution shift indicates that the marginal distribution of $\mathbf{X}$ shifts across environments while the conditional distribution $P(T,Y|\mathbf{X})$ remains unchanged. That is, $\forall e,e' \in \mathcal{E}$, $P^e(T,Y|\mathbf{X})=P^{e'}(T,Y|\mathbf{X})$ and $P^e(\mathbf{X}) \neq P^{e'}(\mathbf{X})$. One naive method to address selection bias and the issue of distribution shift is to combine representation-based methods and stable learning techniques. To this end, we propose a Stable Balanced Representation Learning (SBRL) to estimate HTE across various populations. However, a novel challenge arises, i.e., \textbf{(C3)} the learning of balanced representation and independence-driven weights in SBRL can be interdependent, hence restricting the generalization performance of stable HTE estimation. It should be noticed that current stable learning techniques, designed for typical prediction tasks, learn sample weights by decorrelating the last layer of the network, while balanced representations are required in the first half of the network. Once the balanced representations are updated, adaptive weight modification is necessary, which, however, cannot guarantee feature independence for generalization. 
As a result, prioritizing the optimization of one objective may entail expenses in achieving the other, making it difficult to achieve stable HTE estimation across environments.

% \textbf{Motivations}. 
% As illustrated in Figure~\ref{fig:intro}, there are two main challenges in stable HTE estimation across OOD populations: \textbf{(C1)} selection bias arising from imbalanced treatment assignment, and \textbf{(C2)} distribution shift across various populations. One naive method is to combine representation-based methods and stable learning techniques to address selection bias and the issue of distribution shift. To this end, we propose a Stable Balanced Representation Learning (SBRL) to estimate HTE across various populations. However, a novel challenge arise, i.e., \textbf{(C3)} the learning of balanced representation and independence-driven weights in SBRL can be interdependent, hence restricting the generalization performance of stable HTE estimation. It should be noticed that current stable learning techniques, designed for typical prediction tasks, learn sample weights by decorrelating the last layer of network, while the balanced representations are required in the first half of the network. Once the balanced representations are updated, adaptive weight modification is necessary, which, however, cannot guarantee feature independence for generalization. 
% As a result, prioritizing the optimization of one objective may entail expenses in achieving the other, making it difficult to achieve stable HTE estimation across environments.

To overcome the above challenges, we propose a novel framework named Stable Balanced Representation Learning with Hierarchical-Attention Paradigm (SBRL-HAP) which settles the conflict between balance and independence in a holistic and hierarchical manner.

\section{Methodology}\label{sec:method}

\begin{figure*}[!t]
\centering
\includegraphics[width=0.95\linewidth]{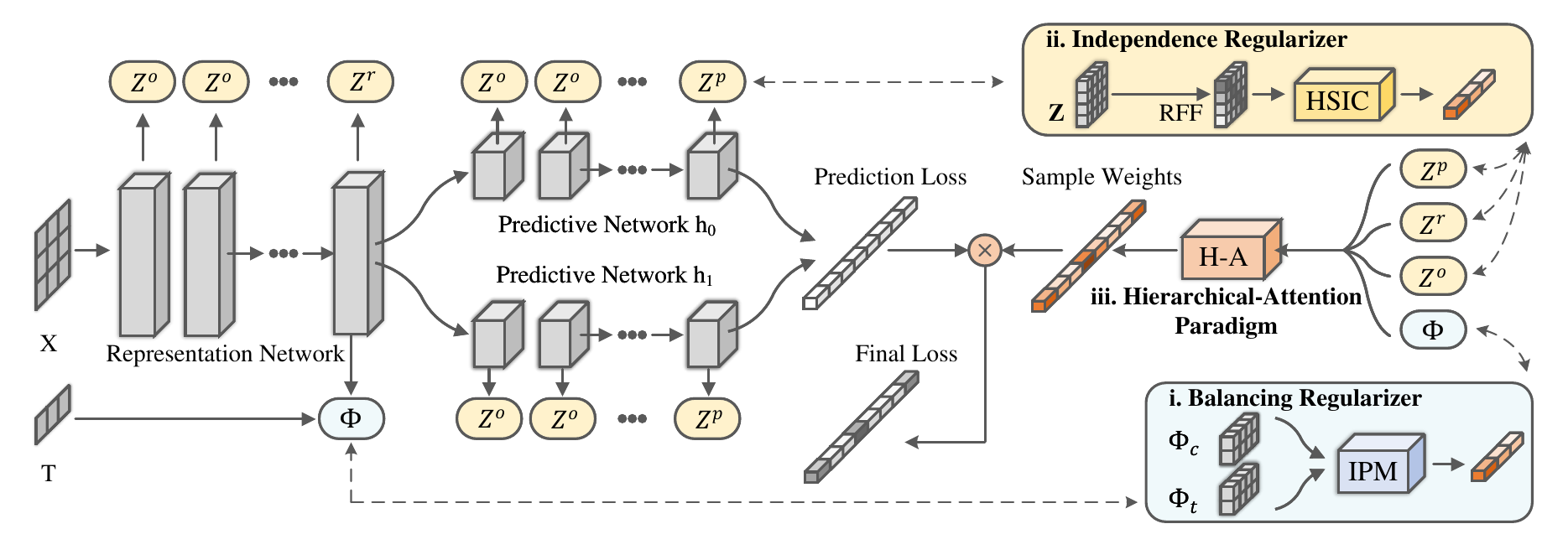}
\caption{The framework of Stable Balanced Representation Learning with Hierarchical-Attention Paradigm (SBRL-HAP). SBRL-HAP consists of three modules: \romannumeral1. Balancing Regularizer restricts IPM for balanced representation, \romannumeral2. Independence Regularizer eliminates feature dependence measured by HSIC-RFF for generalization, and \romannumeral3. Hierarchical-Attention Paradigm decorrelates features comprehensively with a hierarchy for dispelling the interaction between balance and independence. With high flexibility, SBRL-HAP can be plugged into most balanced representation methods by replacing the neural network backbone.}
\label{fig:framework}
\end{figure*}

In this section, we propose SBRL-HAP to stably estimate heterogeneous treatment effects across OOD populations. Firstly, we will present the overall framework of our SBRL-HAP. Subsequently, we will offer a thorough description of three components of SBRL-HAP. Finally, we will demonstrate the end-to-end optimization and training strategies.

Fig.~\ref{fig:framework} depicts the overall architecture of our SBRL-HAP which consists of three components for stable HTE estimation:

\begin{itemize}
\item \textbf{Balancing Regularizer (BR)} employs Integral Probability Metrics (IPM)~\cite{sriperumbudurIntegralProbabilityMetrics2009,mullerIntegralProbabilityMetrics1997} to measure the distribution discrepancy between the treated and control group, and proposes to adopt a model-free method to narrow the distribution discrepancy, so as to eliminate selection bias and obtain balanced representations.
\item \textbf{Independence Regularizer (IR)} learns sample weights to remove non-linear dependencies between features by utilizing the Hilbert-Schmidt Independence Criterion~\cite{grettonKernelStatisticalTest2007} with Random Fourier Features~\cite{stroblApproximateKernelBasedConditional2019}, thereby facilitating the identification of the stable relationships between features and potential outcomes. 
\item \textbf{Hierarchical-Attention Paradigm (HAP)} emphasizes assigning distinct priorities to each neural network layer, in order to achieve comprehensive feature decorrelation with hierarchical attention. Therefore, HAP harmoniously integrates the Balancing Regularizer and Independence Regularizer, effectively resolving the conflict between balance and independence.
\end{itemize}

Next, we will describe each component of our SBRL-HAP model in detail, and then demonstrate the end-to-end optimization and training strategy.

\subsection{Balancing Regularizer}
The Balancing Regularizer is designed to eliminate selection bias and obtain a balanced representation by reducing the distribution discrepancy between different treatment arms with a model-free method.
A typical metric used for measuring the distribution discrepancy is the Integral Probability Metric (IPM)~\cite{sriperumbudurIntegralProbabilityMetrics2009,mullerIntegralProbabilityMetrics1997}, which is formally defined as
\begin{equation}
dist(P_{\Phi_c},P_{\Phi_t}) = \sup\limits_{f\in\mathcal{F}}|\mathbb{E}_{x\sim P_{\Phi_c}}[f(x)]-\mathbb{E}_{x\sim P_{\Phi_t}}[f(x)]|,
\end{equation}
where $P_{\Phi_c}=\{\Phi(\mathbf{x}_i)\}_{i:t_i=0}$ and $P_{\Phi_t}=\{\Phi(\mathbf{x}_i)\}_{i:t_i=1}$ denote the covariate distribution of the control group and the treated group in the representation space $\Phi$, respectively. 
For rich enough function families $\mathcal{F}$, $dist(P_{\Phi_c},P_{\Phi_t})=0 \Rightarrow P_{\Phi_c}=P_{\Phi_t}$ holds~\cite{shalitEstimatingIndividualTreatment2017}.
Most previous works constrain the IPM $dist(P_{\Phi_c},P_{\Phi_t})$ by directly optimizing network parameters~\cite{shalitEstimatingIndividualTreatment2017,johanssonLearningRepresentationsCounterfactual2016}, thereby getting rid of selection bias.
This practice may lead to an overbalanced representation discarding predictive information~\cite{wuLearningDecomposedRepresentations2022}.

Therefore, we propose to adopt the sample reweighting technique to reduce network dependence.
Specifically, our Balancing Regularizer strives to mitigate selection bias by learning a set of sample weights $\mathbf{w}=(w_1,w_2,\dots,w_n)\in \mathbb{R}^n_+$ with minimizing the following balance loss $\mathcal{L}_{\text{B}}$:
\begin{equation}
\min_{\mathbf{w}} \mathcal{L}_{\text{B}} = \sup\limits_{f\in\mathcal{F}}|\mathbb{E}_{x\sim P_{\Phi_c}^\mathbf{w}}[f(x)]-\mathbb{E}_{x\sim P_{\Phi_t}^\mathbf{w}}[f(x)]|,
\end{equation} % = dist^\mathbf{w}(P_{\Phi_c},P_{\Phi_t})
where $P_{\Phi_c}^\mathbf{w}=\{w_i\cdot\Phi(\mathbf{x}_i)\}_{i:t_i=0}$ and $P_{\Phi_t}^\mathbf{w}=\{w_i\cdot\Phi(\mathbf{x}_i)\}_{i:t_i=1}$ denote the weighted probability distributions of covariates in the representation space $\Phi$ with $t=0$ and $t=1$, respectively.

% Selection bias brings about the distribution imbalance between the treated and control group (Top panel in Fig.~\ref{fig:intro}). To get rid of selection bias, most previous work narrows distribution discrepancy by directly optimizing network parameters~\cite{shalitEstimatingIndividualTreatment2017,johanssonLearningRepresentationsCounterfactual2016} or using the propensity score~\cite{hassanpourLearningDisentangledRepresentations2020,hassanpourCounterFactualRegressionImportance2019}.
% The former may lead to an overbalanced representation discarding predictive information, while the latter relies on the correctness of the propensity score model~\cite{wuLearningDecomposedRepresentations2022,hainmuellerEntropyBalancingCausal2012}.

% Therefore, we propose to adopt the sample reweighting technique that reduces model dependence for covariate balancing.
% Specifically, our Balancing Regularizer strives to learn a set of sample weights $\mathbf{w}=(w_1,w_2,\dots,w_n)\in \mathbb{R}^n_+$ to minimize distribution differences between various treatment arms in the representation space $\Phi$.
% In this paper, we apply Integral Probability Metrics (IPM)~\cite{sriperumbudurIntegralProbabilityMetrics2009,mullerIntegralProbabilityMetrics1997} to measure distribution discrepancy, and our Balancing Regularizer optimize sample weights $\mathbf{w}$ by the following balance loss $\mathcal{L}_{\text{B}}$:

\subsection{Independence Regularizer}

The Independence Regularizer aims to eliminate feature dependencies, so as to recognize stable representations against distribution shifts. As stated in previous studies~\cite{cuiStableLearningEstablishes2022,xuTheoreticalAnalysisIndependencedriven2022,zhangDeepStableLearning2021}, the statistical correlation between stable features $\mathbf{X}_S$ and unstable features $\mathbf{X}_V$ is a major cause of model failure under distribution shifts, and thus, independence between variables can lead to more reliable and stable models. When variables are independent, alterations in one variable do not exert any influence on the other variables. Thereby, the relationships between variables and outcomes can be regarded as stable causation, facilitating the superior performance of models across different OOD populations.

The Independence Regularizer employs the Hilbert-Schmidt Independence Criterion with Random Fourier Features to measure the non-linear correlation between two variables.
HSIC is widely utilized to measure the dependency between two random variables by comparing their representations in a Hilbert space~\cite{zhangDeepStableLearning2021,changInformativeSubspaceLearning2017,yaoSCISubspaceLearning2021}:
\begin{equation}
\begin{aligned}
\text{HSIC}(A,B)=\|\mathbf{K}_A-\mathbf{K}_B\|^2_{HS},
\end{aligned}
\end{equation}
where $\mathbf{K}_A=k_A(A,A)$ and $\mathbf{K}_B=k_B(B,B)$ are RBF kernel matrices, and $\|\cdot\|_{HS}$ is the Hilbert-Schmidt norm. If the product $k_Ak_B$ is characteristic, and $\mathbb{E}[k_A(A,A)]<\infty$ and $\mathbb{E}[k_B(B,B)]<\infty$ hold, then $A \perp B$ if and only if $\text{HSIC}(A,B)=0$. However, HSIC involving large-scale kernel matrices is computationally expensive.

Therefore, HSIC with Random Fourier Features (HSIC-RFF) is developed as an approximation technique for HSIC, leading to a notable reduction in time complexity. The function space of Random Fourier Features is:
\begin{equation}
\begin{aligned}
\mathcal{H}_{\text{RFF}}=\{h:x\rightarrow \sqrt{2}\cos{(wx+\varphi)}\},
\end{aligned}
\end{equation}
where $w\sim \mathcal{N}(0,1)$ and $\varphi\sim \mathcal{U}(0,2\pi)$ from normal distribution and the uniform distribution. Then, the statistics of HSIC can be approximated as $\text{HSIC}_{\text{RFF}}$:
\begin{equation}
\begin{aligned}
\text{HSIC}_{\text{RFF}}(A,B) &=\big\|C_{\mathbf{u}(A),\mathbf{v}(B)}\big\|^2_F \\
&=\sum_{i=1}^{n_A}\sum_{j=1}^{n_B}\big|Cov(u_i(A),v_j(B))\big|^2,
\end{aligned}
\end{equation}
where
$\|\cdot\|_F$ is the Frobenius norm, and
$C_{\mathbf{u}(A), \mathbf{v}(B)}\in \mathcal{R}^{n_A\times n_B}$ is the cross-covariance matrix of random Fourier features $\mathbf{u}(A)$ and $\mathbf{u}(B)$ containing entries:
\begin{equation}
\begin{aligned}
&\mathbf{u}(A)=(u_1(A),u_2(A),\dots,u_{n_A}(A)), u_i(A)\in \mathcal{H}_{\text{RFF}},\forall i,\\
&\mathbf{v}(B)=(v_1(B),v_2(B),\dots,v_{n_B}(B)), v_j(B)\in \mathcal{H}_{\text{RFF}},\forall j,
\end{aligned}
\end{equation}
where $n_A$ and $n_B$ denote the number of functions from $\mathcal{H}_{\text{RFF}}$. The accuracy of the statistics $\text{HSIC}_{\text{RFF}}$ increases as the values of $n_A$ and $n_B$, defaulting to 5, become larger.

Motivated by~\cite{zhangDeepStableLearning2021,fanGeneralizingGraphNeural2021a}, our Independence Regularizer coherently optimizes sample weights $\mathbf{w}$ by decorrelating all features in covariates (or its representations) $\mathbf{X}\in \mathbb{R}^{n\times m}$. That is, for any two features $\mathbf{X}_{:,a},\mathbf{X}_{:,b} \in \mathbf{X}$, the weighted statistics $\text{HSIC}_{\text{RFF}}$, denoted by $\text{HSIC}^w_{\text{RFF}}$, should be close to zero.
Formally, for $\forall \mathbf{X}_{:,a},\mathbf{X}_{:,b} \in \mathbf{X}$,
\begin{equation}
\begin{aligned}
& \text{HSIC}^w_{\text{RFF}}(\mathbf{X}_{:,a},\mathbf{X}_{:,b},\mathbf{w})=\big\|C^w_{\mathbf{u}(\mathbf{X}_{:,a}),\mathbf{v}(\mathbf{X}_{:,b})}\big\|^2_F\\
&=\sum_{i=1}^{n_A}\sum_{j=1}^{n_B}\big|Cov(u_i(\mathbf{w}^\top \mathbf{X}_{:,a}),v_j(\mathbf{w}^\top \mathbf{X}_{:,b}))\big|^2\rightarrow 0.
\end{aligned}
\end{equation}

The corresponding loss term can be denoted as:
\begin{equation}\label{eq:LD}
\mathcal{L}_{\text{D}}(\mathbf{X},\mathbf{w})=\sum_{1\leq a\leq b\leq m}\text{HSIC}^w_{\text{RFF}}(\mathbf{X}_{:,a},\mathbf{X}_{:,b},\mathbf{w}).
\end{equation}

Following prior work~\cite{zhangDeepStableLearning2021,fanGeneralizingGraphNeural2021a}, we apply the loss term $\mathcal{L}_{\text{D}}(\cdot,\cdot)$ to the last layer of the neural network $\mathcal{Z}^p$, and thus obtain the independence loss of our Independence Regularizer $\mathcal{L}_{\text{I}}=\mathcal{L}_{\text{D}}(\mathcal{Z}^p,\mathbf{w})$. This is done to ensure that stable representations can establish the most direct mapping to the outcome.

Note that our Balancing Regularizer and Independence Regularizer are designed to handle the issue of selection bias and distribution shifts separately, which are both based on sample reweighting. Therefore, we propose to directly integrate the Balancing Regularizer and the Independence Regularizer to achieve stable and reliable HTE estimation. This approach is named Stable Balanced Representation Learning (SBRL). 
% , and experimental results in Section~\ref{sec:exp} show the performance of our SBRL.
% Because the learning of sample weights $\mathbf{w}$ involves only the representation $\mathcal{Z}^p$ and is regardless of the model structure, the Independence Regularizer can be flexibly employed to handle distribution shifts. This work is one of pioneers to apply the Independence Regularizer into Balanced Representation Learning, addressing the problem of HTE estimation across OOD data (refer to Section~\ref{sec:exp} for experimental results).

\subsection{Hierarchical-Attention Paradigm (HAP)}

Although Balancing Regularizer and Independence Regularizer are able to solve selection bias and distribution shift separately, distribution shift in HTE estimation triggers an extra challenge, i.e., the contradiction between balance and dependence as stated in Section~\ref{sec:prb}. This challenge poses a significant obstacle to reconciling the Balancing Regularizer and Independence Regularizer methods, thereby making it difficult to achieve stable HTE estimates in OOD environments. To address this issue, we propose a Hierarchical-Attention Paradigm to form a coordinated and unified objective function.

The design of HAP stems from the following insight: applying decorrelation solely to the last layer of models, as traditional works suggest~\cite{zhangDeepStableLearning2021,fanGeneralizingGraphNeural2021a}, would induce interaction between the learning of balanced representation and independence-driven weights; one intuitive approach is to uniformly enforce decorrelation for each layer throughout the entire network. However, such indiscriminate constraints may lead to a large value for the independence loss and a relatively small value for the balance loss term, resulting in the disregard of the covariate balancing objective.

% Different from previous work~\cite{zhangDeepStableLearning2021,fanGeneralizingGraphNeural2021a}, which paid attention to the specific last layer of model, our approach suggests holistic decorrelation with priority, so as to achieve end-to-end recognition of stable and balanced representations, resulting in stable HTE estimation across OOD data.
% This practice builds on the following insight: applying decorrelation solely to the last layer of models, as traditional works suggest, would induce interaction between the learning of balanced representation and independence-driven weights; Alternatively, undifferentiated decorrelation for each layer may obscure the focus on balance and independence.
% % This practice builds on the insight that applying decorrelation solely to the last layer of models, as traditional work suggests, would be insufficient to remove feature dependencies within balanced representations; Alternatively, when implemented across all layers of models, it may diffuse the focus.

Consequently, we propose to divide the entire neural network into three priorities: the model's last layer $\mathbf{Z}^p\in \mathbb{R}^{n\times d_p}$ with the first priority, the layer $\mathbf{Z}^r\in \mathbb{R}^{n\times d_r}$ for balanced representations $\Phi$ with the second priority and other fully connected layers $\{\mathbf{Z}^o_i\in \mathbb{R}^{n\times d_o}\}_{i=1}^l$ with the third priority.
Then, besides the loss term $\mathcal{L}_{\text{I}}$ for $\mathbf{Z}^p$, we emphasize the necessity of the loss terms $\mathcal{L}_{\text{D}}(\mathbf{Z}^r,\mathbf{w})$ and $\mathcal{L}_{\text{D}}(\mathbf{Z}^o,\mathbf{w})$ with hierarchical attention for thorough removal of the negative impact of unstable features. 

By integrating Balancing Regularizer and Independence Regularizer with hierarchical attention, the Hierarchical-Attention Paradigm optimizes sample weights $\mathbf{w}$ with the following loss function $\mathcal{L}_{\mathbf{w}}$:
\begin{equation}
\begin{aligned}
\min_\mathbf{w} \mathcal{L}_{\mathbf{w}}=&\alpha\cdot\mathcal{L}_{\text{B}}+\gamma_1\cdot\mathcal{L}_{\text{I}}+\gamma_2\cdot\mathcal{L}_{\text{D}}(\mathbf{Z}^r,\mathbf{w})+\\
&\gamma_3\cdot\sum_{i=1}^l\mathcal{L}_{\text{D}}(\mathbf{Z}^o_i,\mathbf{w})+\mathcal{R}_{\mathbf{w}},
\end{aligned}
\end{equation}
where $\mathcal{R}_{\mathbf{w}}=\frac{1}{n}\sum_{i=1}^n(w_i-1)^2$ avoids all the sample weights to be zero or model only focuses on some samples and ignores others.
Besides, the value of hyper-parameters $\alpha$ and $\{\gamma_1,\gamma_2,\gamma_3\}$ allows us to adjust the sensitivity to selection bias and distribution shift with hierarchical attention.

\renewcommand{\algorithmicrequire}{\textbf{Input:}}
\renewcommand{\algorithmicensure}{\textbf{Output:}}
\begin{algorithm}[t]
    \caption{Stable Balanced Representation Learning with Hierarchical-Attention Paradigm}
    \label{algo}
    \begin{algorithmic}[1]
        \Require Observational dataset $D^e=\{\mathbf{x}_i^e,t_i^e,y_i^{t_i,e}\}^n$ from environment $e$
        \Ensure $\hat{y}^0$, $\hat{y}^1$
        \State Initialize network parameters $\mathbf{W},\mathbf{b}$
        \State Initialize sample weights $\mathbf{w}\gets\{1\}^n$
        \For{$i=0$ \textbf{to} $\mathcal{I}$}
            \State{Calculate loss function $\mathcal{L}_{\text{Y}}^w$ with parameters $\mathbf{W},\mathbf{b}$} and sample weights $\mathbf{w}$
            \State{Update $\mathbf{W},\mathbf{b}$ with gradient descent by fixing $\mathbf{w}$}
            \State{Calculate loss function $\mathcal{L}_{\mathbf{w}}$ with sample weights $\mathbf{w}$}
            \State{Update $\mathbf{w}$ with gradient descent by fixing $\mathbf{W},\mathbf{b}$}
        \EndFor
    \end{algorithmic}
\end{algorithm}

\subsection{Optimization and Training Procedure}
By optimizing the loss function $\mathcal{L}_{\mathbf{w}}$, we can acquire the optimal sample weights $\mathbf{w}^*$ to guide the deep neural networks to achieve stable HTE estimation across OOD data. Note that our SBRL-HAP learns sample weights regardless of the model structure; hence, it is applicable to the backbone of nearly all balanced representation methods. We take the backbone of the most classic balanced representation algorithm, i.e., Counterfactual Regressor (CFR) \cite{shalitEstimatingIndividualTreatment2017}, as an example, to illustrate our end-to-end training process.

The backbone of CFR contains two sub-modules, i.e., a shared representation network ($\Phi(\mathbf{x}_i)$) for representation extraction and multi-head predictive networks ($h_{t_i}(\Phi(\mathbf{x}_i)$) for potential outcome prediction. 

The representation network is expected to provide a balanced representation $\Phi(\mathbf{X})$, so as to remove distribution discrepancies between the treated group $\{\Phi(\mathbf{x}_i)\}_{i:t_i=1}$ and the control group $\{\Phi(\mathbf{x}_i)\}_{i:t_i=0}$. Then, in HTE estimation, to avoid the treatment information being dominated by the high-dimensional covariates, the two-head networks $h_{t=0}(\Phi)$ and $h_{t=1}(\Phi)$ are adopted to predict outcomes in control and treated groups, with the prediction loss $\mathcal{L}_Y$:
\begin{equation}\label{eq:loss_y}
\min_{h_0,h_1} \mathcal{L}_{\text{Y}} = \frac{1}{n}\sum_{i=1}^nl(h_{t_i}(\Phi(\mathbf{x}_i)),y_i^{t_i}) + \mathcal{R}_{l_2},
\end{equation}
where $\mathcal{R}_{l_2}$ is $l_2$ regularization for $h$, and $l(\cdot,\cdot)$ encodes the loss function, i.e., mean squared loss (MSE) for continuous outcome, and cross-entropy error for binary outcome.

To guide the above neural networks to achieve stable and unbiased prediction, we propose to plug our SBRL-HAP module in Equation~(\ref{eq:loss_y}) by
\begin{equation}\label{eq:loss_yw}
\min_{h_0,h_1} \mathcal{L}_{\text{Y}}^w = \frac{1}{n}\sum_{i=1}^nw_i\cdot l(h_{t_i}(\Phi(\mathbf{x}_i)),y_i^{t_i})+\mathcal{R}_{l_2},
\end{equation}
where $\{w_i\}_{i=1}^n\in \mathbf{w}^*$ are the optimal sample weights learnt with the loss function $\mathcal{L}_{\mathbf{w}}$.

Ultimately, we adopt an alternating training strategy to iteratively optimize the loss function $\mathcal{L}_{\text{Y}}^w $ for heterogeneous outcome prediction and the loss function $\mathcal{L}_{\mathbf{w}}$ for stable and balanced representations.
Algorithm~\ref{algo} illustrates the details of the pseudo-code of our SBRL-HAP.

\section{Experiments}\label{sec:exp}
% Table generated by Excel2LaTeX from sheet 'Syn_8'
\begin{table*}[htbp]
  \centering
  \caption{The results (mean$\pm$std) of treatment effect estimation on synthetic data $\text{Syn}\_8\_8\_8\_2$ with different bias rate $\rho$.}
  \begingroup %这一行要加！！！！！！
  \setlength{\tabcolsep}{5pt} % 调整列间距  Default value: 6pt
  \renewcommand{\arraystretch}{0.8} % 调整行间距  Default value: 1
    \begin{threeparttable}
    \begin{tabular}{lcccccccc}
    \toprule
    Metric & \multicolumn{8}{c}{\boldmath{}\textbf{PEHE (Mean$\pm$Std)}\unboldmath{}} \\
    \midrule
    Bias Rate & $\bm{\rho=-3}$ & $\bm{\rho=-2.5}$ & $\bm{\rho=-1.5}$ & $\bm{\rho=-1.3}$ & $\bm{\rho=1.3}$ & $\bm{\rho=1.5}$ & $\bm{\rho=2.5}^*$ & $\bm{\rho=3}$ \\
    \midrule
    TARNet & 0.565$\pm$0.009 & 0.558$\pm$0.007 & 0.567$\pm$0.003 & 0.559$\pm$0.006 & 0.461$\pm$0.003 & 0.420$\pm$0.005 & 0.363$\pm$0.003 & 0.358$\pm$0.003 \\
    +SBRL & 0.474$\pm$0.008 & 0.459$\pm$0.008 & 0.492$\pm$0.007 & 0.489$\pm$0.009 & \boldmath{}\textbf{0.410$\pm$0.007}\unboldmath{} & \boldmath{}\textbf{0.377$\pm$0.004}\unboldmath{} & \boldmath{}\textbf{0.341$\pm$0.004}\unboldmath{} & \boldmath{}\textbf{0.332$\pm$0.004}\unboldmath{} \\
    +SBRL-HAP & \boldmath{}\textbf{0.440$\pm$0.005}\unboldmath{} & \boldmath{}\textbf{0.435$\pm$0.007}\unboldmath{} & \boldmath{}\textbf{0.442$\pm$0.008}\unboldmath{} & \boldmath{}\textbf{0.462$\pm$0.004}\unboldmath{} & 0.444$\pm$0.005 & 0.421$\pm$0.006 & 0.404$\pm$0.006 & 0.407$\pm$0.005 \\
    \midrule
    CFR & 0.559$\pm$0.009 & 0.552$\pm$0.007 & 0.563$\pm$0.003 & 0.555$\pm$0.006 & 0.459$\pm$0.003 & 0.418$\pm$0.005 & 0.363$\pm$0.003 & 0.357$\pm$0.003 \\
    +SBRL & 0.475$\pm$0.008 & 0.460$\pm$0.008 & 0.492$\pm$0.007 & 0.490$\pm$0.009 & 0.410$\pm$0.007 & 0.378$\pm$0.004 & \boldmath{}\textbf{0.341$\pm$0.004}\unboldmath{} & \boldmath{}\textbf{0.332$\pm$0.004}\unboldmath{} \\
    +SBRL+HAP & \boldmath{}\textbf{0.419$\pm$0.005}\unboldmath{} & \boldmath{}\textbf{0.412$\pm$0.005}\unboldmath{} & \boldmath{}\textbf{0.429$\pm$0.004}\unboldmath{} & \boldmath{}\textbf{0.433$\pm$0.005}\unboldmath{} & \boldmath{}\textbf{0.401$\pm$0.007}\unboldmath{} & \boldmath{}\textbf{0.374$\pm$0.006}\unboldmath{} & 0.354$\pm$0.005 & 0.352$\pm$0.005 \\
    \midrule
    DeRCFR & 0.431$\pm$0.007 & 0.439$\pm$0.009 & 0.449$\pm$0.007 & 0.455$\pm$0.008 & 0.376$\pm$0.005 & 0.338$\pm$0.005 & 0.311$\pm$0.004 & 0.306$\pm$0.005 \\
    +SBRL & 0.431$\pm$0.005 & 0.429$\pm$0.007 & 0.441$\pm$0.004 & 0.446$\pm$0.007 & 0.371$\pm$0.006 & 0.335$\pm$0.006 & 0.301$\pm$0.006 & \boldmath{}\textbf{0.293$\pm$0.002}\unboldmath{} \\
    +SBRL-HAP & \boldmath{}\textbf{0.350$\pm$0.006}\unboldmath{} & \boldmath{}\textbf{0.353$\pm$0.009}\unboldmath{} & \boldmath{}\textbf{0.373$\pm$0.006}\unboldmath{} & \boldmath{}\textbf{0.374$\pm$0.009}\unboldmath{} & \boldmath{}\textbf{0.340$\pm$0.006}\unboldmath{} & \boldmath{}\textbf{0.312$\pm$0.006}\unboldmath{} & \boldmath{}\textbf{0.295$\pm$0.006}\unboldmath{} & 0.295$\pm$0.006 \\
    \midrule
    Improvement & $25.0\%\uparrow$ & $25.4\%\uparrow$ & $23.8\%\uparrow$ & $22.0\%\uparrow$ & $12.6\%\uparrow$ & $10.5\%\uparrow$ & $5.1\%\uparrow$ & $3.6\%\uparrow$ \\
    \midrule
    \midrule
    Metric & \multicolumn{8}{c}{$\bm{\epsilon_{\text{ATE}}}$ \textbf{(Mean$\pm$Std)}} \\
    \midrule
    Bias Rate & $\bm{\rho=-3}$ & $\bm{\rho=-2.5}$ & $\bm{\rho=-1.5}$ & $\bm{\rho=-1.3}$ & $\bm{\rho=1.3}$ & $\bm{\rho=1.5}$ & $\bm{\rho=2.5}^*$ & $\bm{\rho=3}$ \\
    \midrule
    TARNet & \boldmath{}\textbf{0.019$\pm$0.006}\unboldmath{} & 0.032$\pm$0.008 & 0.012$\pm$0.004 & 0.015$\pm$0.005 & \boldmath{}\textbf{0.021$\pm$0.008}\unboldmath{} & 0.021$\pm$0.008 & 0.018$\pm$0.006 & 0.021$\pm$0.007 \\
    +SBRL & 0.029$\pm$0.005 & 0.040$\pm$0.006 & 0.027$\pm$0.004 & 0.026$\pm$0.005 & 0.029$\pm$0.011 & 0.020$\pm$0.006 & 0.026$\pm$0.006 & 0.029$\pm$0.008 \\
    +SBRL-HAP & 0.021$\pm$0.006 & \boldmath{}\textbf{0.025$\pm$0.009}\unboldmath{} & \boldmath{}\textbf{0.012$\pm$0.004}\unboldmath{} & \boldmath{}\textbf{0.015$\pm$0.005}\unboldmath{} & 0.023$\pm$0.008 & \boldmath{}\textbf{0.019$\pm$0.008}\unboldmath{} & \boldmath{}\textbf{0.017$\pm$0.007}\unboldmath{} & \boldmath{}\textbf{0.021$\pm$0.007}\unboldmath{} \\
    \midrule
    CFR & \boldmath{}\textbf{0.018$\pm$0.006}\unboldmath{} & 0.032$\pm$0.008 & \boldmath{}\textbf{0.012$\pm$0.004}\unboldmath{} & 0.014$\pm$0.004 & \boldmath{}\textbf{0.021$\pm$0.008}\unboldmath{} & 0.020$\pm$0.008 & 0.018$\pm$0.006 & 0.021$\pm$0.007 \\
    +SBRL & 0.029$\pm$0.005 & 0.040$\pm$0.006 & 0.028$\pm$0.004 & 0.026$\pm$0.005 & 0.030$\pm$0.011 & 0.021$\pm$0.006 & 0.027$\pm$0.006 & 0.029$\pm$0.008 \\
    +SBRL-HAP & 0.019$\pm$0.006 & \boldmath{}\textbf{0.024$\pm$0.009}\unboldmath{} & 0.015$\pm$0.005 & \boldmath{}\textbf{0.013$\pm$0.004}\unboldmath{} & 0.024$\pm$0.008 & \boldmath{}\textbf{0.018$\pm$0.006}\unboldmath{} & \boldmath{}\textbf{0.013$\pm$0.006}\unboldmath{} & \boldmath{}\textbf{0.015$\pm$0.007}\unboldmath{} \\
    \midrule
    DeRCFR & 0.017$\pm$0.006 & \boldmath{}\textbf{0.021$\pm$0.007}\unboldmath{} & 0.014$\pm$0.004 & 0.020$\pm$0.005 & \boldmath{}\textbf{0.021$\pm$0.008}\unboldmath{} & 0.020$\pm$0.007 & 0.019$\pm$0.006 & 0.021$\pm$0.006 \\
    +SBRL & 0.021$\pm$0.007 & 0.033$\pm$0.005 & 0.024$\pm$0.005 & 0.028$\pm$0.006 & 0.027$\pm$0.011 & 0.018$\pm$0.005 & 0.022$\pm$0.007 & 0.029$\pm$0.008 \\
    +SBRL-HAP & \boldmath{}\textbf{0.013$\pm$0.003}\unboldmath{} & 0.023$\pm$0.008 & \boldmath{}\textbf{0.013$\pm$0.005}\unboldmath{} & \boldmath{}\textbf{0.015$\pm$0.005}\unboldmath{} & 0.022$\pm$0.009 & \boldmath{}\textbf{0.013$\pm$0.005}\unboldmath{} & \boldmath{}\textbf{0.019$\pm$0.007}\unboldmath{} & \boldmath{}\textbf{0.021$\pm$0.008}\unboldmath{} \\
    \midrule
    Improvement & $23.5\%\uparrow$ & $25.0\%\uparrow$ & $7.1\%\uparrow$ & $25.0\%\uparrow$ & $4.8\%\downarrow$ & $35.0\%\uparrow$ & $27.8\%\uparrow$ & $28.6\%\uparrow$ \\
    \bottomrule
    \end{tabular}%
    \begin{tablenotes}   
        \footnotesize      
        \item[*] In this paper, we utilize synthetic data with $\rho=2.5$ as the training population. The testing data with $\rho=2.5$ can be regarded as the In-Distribution Population. As the parameter $\rho$ increases, the difference in distribution between the testing and training datasets also increases.
        \end{tablenotes}
    \end{threeparttable}
  \endgroup
  \label{tb:Syn_i8_c8_a8_v2}%
\end{table*}%
\subsection{Baselines}

In this paper, we propose two model-agnostic frameworks, \textbf{SBRL} and \textbf{SBRL-HAP}\footnote{https://github.com/superpig99/SBRL-HAP}, for estimating heterogeneous treatment effects across out-of-distribution environments. {SBRL} can be regarded as an ablation study of {SBRL-HAP} without HAP. In these frameworks, most existing representation balancing methods can be incorporated as backbones, because our methods only introduce BR, IR, and HAP as additional regularizers to constrain representation learning, without being tied to specific models.
Below, to demonstrate the performance of our {SBRL} and {SBRL-HAP} in improving heterogeneous treatment effect estimation across OOD populations, we compare them to baselines and describe how we can combine {SBRL} and {SBRL-HAP} with each method:

% we compare SBRL-HAP with three vanilla balanced representation methods.

% In this paper, we propose a model-agnostic framework called \textbf{SBRL-HAP} for estimating heterogeneous treatment effects across out-of-distribution (OOD) environments. In this framework, all existing representation balancing methods can be incorporated as backbones because our method only introduces BR, IR, and HAP as additional regularizers to constrain representation learning, without being tied to specific models. 

%To demonstrate the performance of our \textbf{SBRL} and \textbf{SBRL-HAP} in improving heterogeneous treatment effect across OOD environments, we compare SBRL-HAP with three vanilla balanced representation methods.
\begin{itemize}
    \item \textbf{TARNet}~\cite{shalitEstimatingIndividualTreatment2017} is a treatment-agnostic representation network algorithm with a shared representation network, which uses a two-head predictive network to predict the factual treated outcome and control outcome, separately. Since TARNet does not include balance regularization, we only incorporate Independence Regularize into TARNet as \textbf{TARNet+SBRL}. \textbf{TARNet+SBRL-HAP} achieves comprehensive feature decorrelation with hierarchical attention by Hierarchical-Attention Paradigm.
    \item \textbf{CFR}~\cite{shalitEstimatingIndividualTreatment2017,johanssonLearningRepresentationsCounterfactual2016} employs IPM to measure the distribution distance between the treated and control groups, and learns a balanced representation by minimizing IPM to eliminate selection bias. By incorporating Balancing Regularization and Independence Regularization into CFR, we refer to it as \textbf{CFR+SBRL}. Furthermore, \textbf{CFR+SBRL-HAP} employs the Hierarchical-Attention Paradigm for comprehensive feature decorrelation through hierarchical attention mechanisms.
    \item \textbf{DeR-CFR}~\cite{wuLearningDecomposedRepresentations2022} further considers confounder separation by learning representations for instrumental variables, confounding variables, and adjustment variables respectively. This enables a more precise evaluation of heterogeneous treatment effects. When incorporating the SBRL framework, we refer to it as \textbf{DeR-CFR+SBRL}. Additionally, when incorporating it into SBRL-HAP framework, we call it \textbf{DeR-CFR+SBRL-HAP}.
\end{itemize}

The aforementioned three baselines are the most classic solutions to the traditional HTE estimation problem within in-distribution populations, we use them as \textbf{Vanilla} models to compare them with \textbf{+SBRL} and \textbf{+SBRL-HAP} models. 
Other balanced representation methods, such as 
RCFR~\cite{johanssonLearningWeightedRepresentations2018}, CFR-ISW~\cite{hassanpourCounterFactualRegressionImportance2019}, SITE~\cite{yaoRepresentationLearningTreatment2018}, and DR-CFR~\cite{hassanpourLearningDisentangledRepresentations2020}, are built upon these baselines, and these methods have not exceeded the performance of DeR-CFR~\cite{wuLearningDecomposedRepresentations2022}. Consequently, we only combine {SBRL} and {SBRL-HAP} with TARNet, CFR, and DeR-CFR to study the performance of our methods in estimating HTE across OOD populations.

% and the performances of these methods still can not exceed DeR-CFR~\cite{wuLearningDecomposedRepresentations2022}. Thus, we only combine {SBRL} and {SBRL-HAP} with these three methods, and study the performance of HTE estimation across OOD populations by comparing with them.

% and thus it is sufficient for our plug-in SBRL-HAP framework to showcase our performance in the issue of HTE estimation across OOD populations by comparing with them. 

% 
\subsection{Metrics}

Following previous work~\cite{shalitEstimatingIndividualTreatment2017,wuLearningDecomposedRepresentations2022}, we adopt the Precision in Estimation of Heterogeneous Effect (PEHE)~\cite{hillBayesianNonparametricModeling2011} and the bias of ATE prediction ($\epsilon_{\text{ATE}}$) to evaluate the individual-level and population-level performance respectively, where $\text{PEHE}=\sqrt{\frac{1}{n}\sum_{i=1}^n((\hat{y}_i^1-\hat{y}_i^0)-(y_i^1-y_i^0))^2}$ and $\epsilon_{\text{ATE}}=|ATE-\hat{ATE}|$. Smaller values of these two metrics indicate better model performance. 

Besides, popular evaluation metrics for prediction tasks, such as $F_1$ Score~\cite{kunschJackknifeBootstrapGeneral1989}, are also adopted to assist in evaluating the model performance.
We utilize the average and stability of error~\cite{kuangStablePredictionModel2020} to evaluate the generalization performance. For example, the average of $F_1$ Score is defined as
$\bar{F_{1}}=\frac{1}{|\mathcal{E}|}\sum_{e\in \mathcal{E}}F_1^e$, and the stability of $F_1$ Score is
$F_1^{std}=\frac{1}{|\mathcal{E}|}\sum_{e\in \mathcal{E}}(F_1^e-\bar{F_{1}})^2$. Lower values of these two indicators mean better model stability.
\subsection{Experimental Settings}
In the experiment, we utilize ELU as the non-linear activation function and adopt the Adam optimizer to train all methods. We set the maximum number of iterations to $3000$. Besides, we apply an exponentially decaying learning rate~\cite{duchiAdaptiveSubgradientMethods2011} and report the best-evaluated iterate with early stopping. We first identify the optimal hyper-parameters for all baseline algorithms by optimizing hyper-parameters with trails on random search~\cite{bergstraRandomSearchHyperparameter2012}. Then, with the fixed basic hyper-parameters, we conduct a random search for the hyper-parameters $\{\gamma_1,\gamma_2,\gamma_3\}$ of HSIC losses in the scope $\{0.0001,0.001,0.01,0.1,1,10,100\}$ to optimize our model.

\subsection{Experiments on Synthetic Data}

\subsubsection{Datasets}
To simulate complex real-world scenarios, the synthetic data used in our study incorporates several key factors:
1) The observed covariates include not only confounding variables but also other relevant factors; 2) The imbalanced treatment assignment would introduce selection bias, reflecting the inherent biases that exist in observational studies; and 3) The synthetic data also incorporates distribution shifts that occur across different environments or populations.
We generate synthetic data using the following process.
%The following is the data generation process.

\textbf{Covariates generation.} 
We generate covariates from a multi-variable normal distribution, i.e., $X_1,X_2,\ldots,X_m \sim \mathcal{N}(0,1)$, where $m=m_I+m_C+m_A+m_V$ denotes the dimension of covariates, and $\{m_I, m_C, m_A, m_V\}$ denote the dimensions of instruments $I$, confounders $C$, adjustments $A$ and noise $V$, respectively.
For generality, we design two settings of variable dimensions $\{m_I,m_C,m_A,m_V\}=\{8,8,8,2\}$ or $\{16,16,16,2\}$ with the sample size $n=10000$, and denoted different setting as $\text{Syn}\_m_I\_m_C\_m_A\_m_V$.

\textbf{Treatments generation.} We produce treatment $t \sim \mathcal{B}(\frac{1}{1+e^{-z}})$, where $z=\frac{1}{10}\theta_t \times X_{IC}+\xi$, $X_{IC}$ denotes the covariates that belong to $I$ and $C$, and $\theta_t\sim \mathcal{U}((8,16)^{m_I+m_C})$. 

\textbf{Outcomes generation.}
Two potential outcomes are generated as follows: $Y^0=\text{sign}(\max(0,z^0-\bar{z^0}))$ and $Y^1=\text{sign}(\max(0,z^1-\bar{z^1}))$, where $z^0=\frac{1}{10}\frac{\theta_{y^0}\times X_{CA}}{m_C+m_A}$, $z^1=\frac{1}{10}\frac{\theta_{y^1}\times X^2_{CA}}{m_C+m_A}$, and $\theta_{y_0},\theta_{y_1}\sim \mathcal{U}((8,16)^{m_C+m_A})$ and $\xi \sim \mathcal{N}(0,1)$. The observed outcome is $Y=TY^1+(1-T)Y^0$.

Finally, to simulate the distribution shift, we generate different covariate distributions by biased sampling.
For each sample, we select it with probability $\Pr=\prod_{X_i\in X_V}|\rho|^{-10*D_i}$, where $D_i=|Y^1-Y^0-\text{sign}(\rho)*X_i|$.
If $\rho>0$, $\text{sign}(\rho)=1$; otherwise, $\text{sign}(\rho)=-1$. 
We generate different data distributions by altering 
the bias rate $\rho\in\{-3.0,-2.5,-1.5,-1.3,1.3,1.5,2.5,3.0\}$, where
$\rho>1$ implies the positive correlation between outcome $Y$ and unstable features $X_V$, and $\rho<-1$ implies the negative correlation.
The higher $|\rho|$ is, the stronger correlation between $Y$ and $X_V$.
Therefore, different values of $\rho$ refer to different environments.
To evaluate the generalization of our SBRL and SBRL-HAP frameworks, we use the generated data with $\rho=2.5$ as default training data, and use the data with different $\rho\in\{-3.0,-2.5,-1.5,-1.3,1.3,1.5,2.5,3.0\}$ as testing data with different environments.

\begin{figure*}[htbp]
  \centering
  \begin{minipage}[b]{0.32\linewidth}
    \centering
    \includegraphics[width=\linewidth]{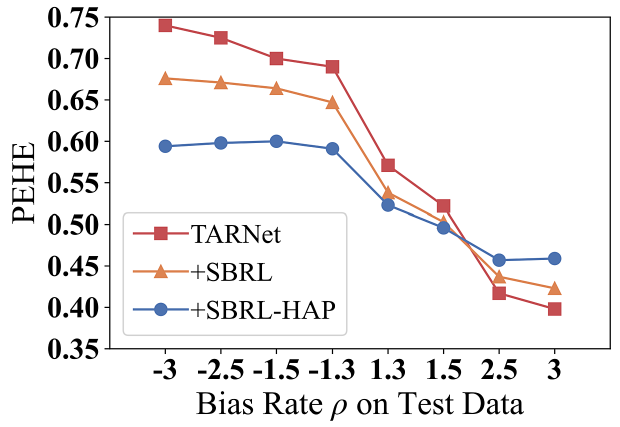}
  \end{minipage}
  \begin{minipage}[b]{0.32\linewidth}
    \centering
    \includegraphics[width=\linewidth]{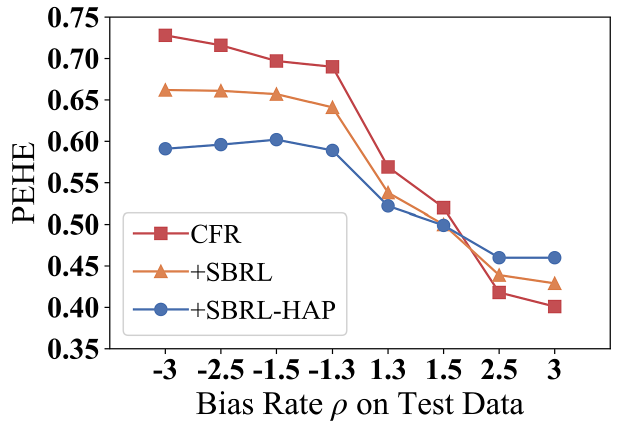}
  \end{minipage}
  \begin{minipage}[b]{0.32\linewidth}
    \centering
    \includegraphics[width=\linewidth]{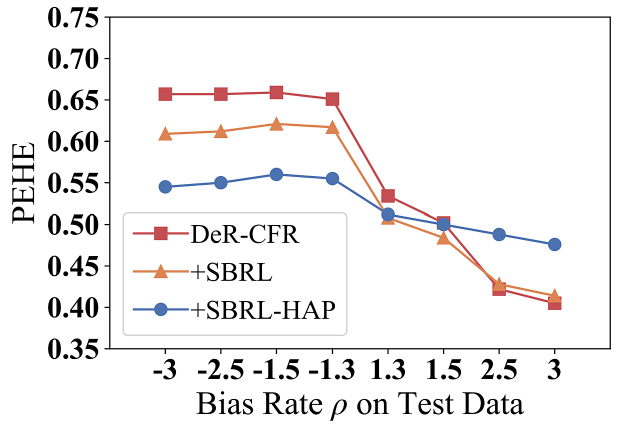}
  \end{minipage}
  \caption{Results of PEHE on synthetic data $\text{Syn}\_16\_16\_16\_2$ with different bias rate $\rho$ for the testing set. All models are trained with $\rho=2.5$.}
  \label{fig:Exp_Syn_16_Pehe}
\end{figure*}

\begin{figure*}[htbp]
  \centering
  \subfigure[$F_1$ scores for factual outcomes.]{
    \includegraphics[width=0.385\textwidth]{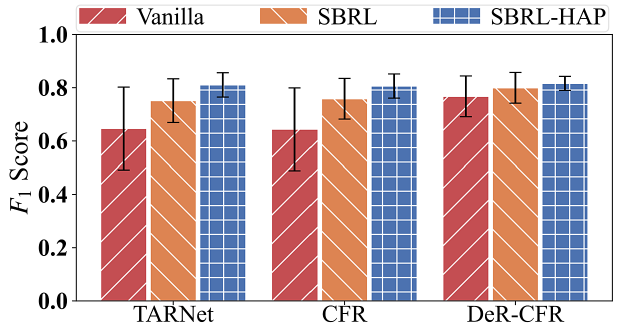}
    \label{fig:subfiga}
  }
  \hspace{0.05\textwidth}
  \subfigure[$F_1$ scores for counterfactual outcomes.]{
    \includegraphics[width=0.385\textwidth]{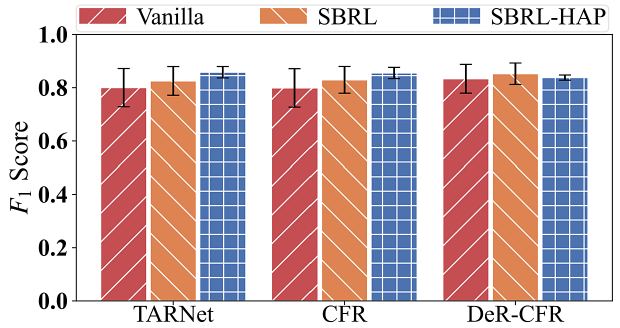}
    \label{fig:subfigb}
  }
  \caption{Results of $F_1$ scores on synthetic data $\text{Syn}\_16\_16\_16\_2$ with different bias rate $\rho$ for the testing set. All models are trained with $\rho=2.5$.}
  \label{fig:Exp_Syn_16_F1score}
\end{figure*}

\subsubsection{Results of treatment effect estimation}
Results of treatment effect estimation on synthetic data are shown in Table~\ref{tb:Syn_i8_c8_a8_v2} and Fig.~\ref{fig:Exp_Syn_16_Pehe}. 
Table~\ref{tb:Syn_i8_c8_a8_v2} reveals that both SBRL and SBRL-HAP effectively boost the stability of ITE estimations across diverse OOD data, while presenting a comparable performance in ATE evaluation compared to the vanilla methods.
According to Table~\ref{tb:Syn_i8_c8_a8_v2}, with the increasing distribution discrepancy between the testing set and the training set, the error metric PEHE of all methods gets worse. Our methods, however, show success in counteracting this performance degradation, and exhibit a more obvious improvement as the bias rate $\rho$ decreases from $2.5$ to $-3$, resulting in the maximum reduction of PEHE from $5.1\%$ to $25\%$.
To validate the robustness of our method for high-dimensional data, we report the results of effect estimation on $\text{Syn}\_16\_16\_16\_2$ data. Fig.~\ref{fig:Exp_Syn_16_Pehe} depicts the excellent performance of our method on high-dimensional data. From results on $\text{Syn}\_16\_16\_16\_2$ data, we have following observations and analysis:
% From results on $\text{Syn}\_8\_8\_8\_2$ data (Table~\ref{tb:Syn_i8_c8_a8_v2}), 
% Fig.~\ref{fig:Exp_Syn_16_Pehe} presents results of treatment effect estimation on synthetic test data $\text{Syn}\_16\_16\_16\_2$ with different bias rate $\rho$. From Fig.~\ref{fig:Exp_Syn_16_Pehe}, we have following observations and analysis:

\begin{itemize}
\item Three baselines fail to handle the problem of HTE estimation accompanied by distribution shifts. On the testing data with $\rho=2.5$, which shares the same distribution as the training test, PEHE is $0.417$, $0.418$, and $0.422$ for TARNet, CFR, and DeR-CFR. However, the performance of the baseline methods degrades gradually as the distribution gap between the testing data and the training data increases (i.e., as $\rho$ decreases). For instance, on the testing data with $\rho=-3$, PEHE of TARNet, CFR, and DeR-CFR worsens to $0.740$, $0.728$, and $0.625$, with the performance decrease\footnote{Performance decrease in OOD testing datasets is calculated by:\\ $\text{Decrease}=(\text{PEHE}_{\{\rho=-3\}}-\text{PEHE}_{\{\rho=2.5\}})/\text{PEHE}_{\{\rho=2.5\}}$.} of $77\%$, $74\%$, and $56\%$, respectively. Such performance degradation of baseline methods is anticipated, as they erroneously capture the spurious correlation between unstable variables $X_V$ and the target outcome $Y$.
\item Compared to other baselines, DeR-CFR exhibits superior resistance to distribution shift, whose performance degradation is about $20\%$ less than TARNet and CFR. This is attributed to DeR-CFR's confounder separation, which orthogonalizes confounding, instrumental, and adjustment variables. It indicates that decorrelating variables is beneficial in learning genuine and stable relationships.
\item Both SBRL and SBRL-HAP achieve more stable HTE estimation across various OOD data. With distribution shifts (i.e., $\rho$ shifts from $2.5$ to $-3$), the PEHE of DeR-CFR+SBRL varies from $0.428$ to $0.609$, indicating a $42\%$ drop. By combining SRBL-HAP, the PEHE of DeR-CFR changes from $0.488$ to $0.545$, only reduced by $11\%$. However, the PEHE of origin DeR-CFR declines by $56\%$. This percentage demonstrates that our algorithm is more stable and the results are more robust in the OOD testing data. Besides, our algorithm exceeds all baselines on each OOD testing data (i.e., $\rho\in [-3, 1.3]$). For example, by combining our SBRL-HAP, the PEHE of DeR-CFR under $\rho=-3$ reduces from $0.657$ to $0.545$, with a $21\%$ performance improvement.
It is because our algorithm resolves the conflict between balance and independence by hierarchical decorrelation, obtaining stable and balanced representations. Hence, our algorithm can improve the stability of HTE estimation.
% \item SBRL improves the generalization ability of three baseline methods. PEHE under $\rho=-3$ reduces to $0.676$, $0.662$ and $0.609$ for TARNet+SBRL, CFR+SBRL and DER-CFR+SBRL, alleviating the performance degradation by $13\%\sim 23\%$.
% Despite some performance improvements, results on our methods show that directly integrating stable learning techniques is insufficient, as this practice does not reconcile the tension between balance and independence, and keeps the adverse impact of unstable variables remaining in balanced representations.
% \item Compared to vanilla and SBRL models, our SBRL-HAP achieves more stable heterogeneous treatment effect estimation across various OOD data. By combining our SBRL-HAP, PEHE of TARNet, CFR and DeR-CFR under $\rho=-3$ reduces to $0.594$, $0.591$ and $0.545$, narrowing the performance degradation \textcolor{blue}{by $44\%\sim47\%$}.
% % from $56\%\sim77\%$ to $11\%\sim29\%$.
% By hierarchical decorrelation, our SBRL-HAP effectively resolves the conflict between balance, and obtains stable and balanced representations. Therefore, our method enables to improve the stability of HTE estimation across distinct OOD populations.
\item Our approach outperforms baselines on OOD data ($\rho<2.5$) but performs worse on ID data ($\rho\geq 2.5$), which aligns with prior observations~\cite{zhangFreeLunchDomain2023,tanProvablyInvariantLearning2023,kruegerOutofDistributionGeneralizationRisk2021a,zhangDomainSpecificRiskMinimization2023,zhouSparseInvariantRisk2022,kuangStablePredictionModel2020,zhangMAPBalancedGeneralization2023}. It is because unstable features tend to contribute to better inference in ID data~\cite{kuangStablePredictionModel2020}; however, our algorithm mitigates the influence of these features to prevent instability when estimating in OOD data.
\end{itemize}

Furthermore, Fig.~\ref{fig:Exp_Syn_16_F1score} demonstrates that our method outperforms the other methods in stably predicting factual and counterfactual outcomes, as measured by $F_1$ scores with mean and standard
deviation (std) across all test sets. Especially, our SBRL-HAP reduces the std of $F_1$ scores from $0.058$ to $0.026$ for factual outcomes and from $0.040$ to $0.009$ for counterfactual outcomes, compared to the best baseline (i.e., DeR-CFR). Consequently, our method can significantly improve the stability of HTE estimation.

% \begin{figure}[tbp]
%   \centering
%   \begin{minipage}{0.325\linewidth}
%     \centering
%     \includesvg[width=\linewidth]{figures/Exp_Heatmap_cfr.svg}
%   \end{minipage}
%   \begin{minipage}{0.325\linewidth}
%     \centering
%     \includesvg[width=\linewidth]{figures/Exp_Heatmap_cfr_ir.svg}
%   \end{minipage}
%   \begin{minipage}{0.325\linewidth}
%     \centering
%     \includesvg[width=\linewidth]{figures/Exp_Heatmap_cfr_ours.svg}
%   \end{minipage}
%   \caption{Nonlinear dependence among features in the balanced representation. As shown, the feature dependence is reduced by our SBRL, and further decreased by incorporating HAP.}
%   \label{fig:Exp_Heatmap}
% \end{figure}

\begin{figure}[tbp]
\centering
\includegraphics[width=\linewidth]{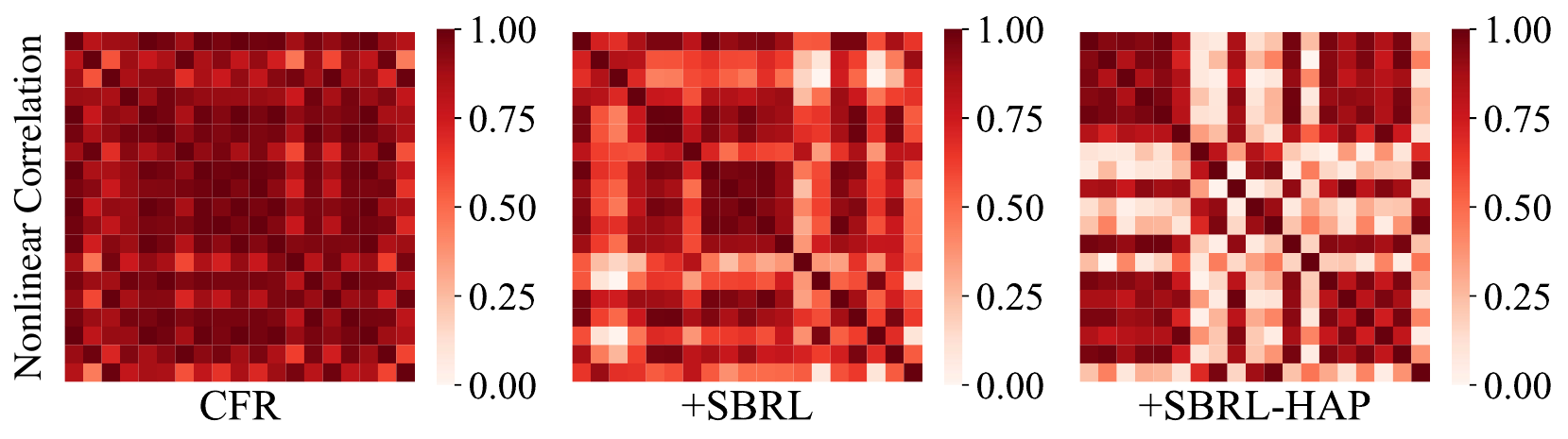}
\caption{Nonlinear correlation among features in the balanced representation. As shown, the feature correlation is reduced by our SBRL, and further decreased by incorporating HAP.}
\label{fig:Exp_Heatmap}
\end{figure}
\setlength{\intextsep}{6pt} 

\subsubsection{Decorrelation Performance}

We demonstrate the nonlinear correlation between features in the balanced representation $\Phi$ to illustrate the effectiveness of our method in mitigating the conflict between balance and independence.
Specifically, we randomly sample 25-dimension variables from the balanced representation learned by CFR, CFR+SBRL, and CFR+SBRL-HAP on data $\text{Syn}\_16\_16\_16\_2$, and compute $\text{HSIC}_{\text{RFF}}$ between each pair of variables.
As shown in Fig.~\ref{fig:Exp_Heatmap}, the balanced representation obtained from CFR exhibits strong correlation between features, with average $\text{HSIC}_{\text{RFF}}=0.85$, while direct integration of representation balancing and stable training techniques (i.e., CFR+SBRL) reduces the average $\text{HSIC}_{\text{RFF}}$ to $0.64$. Notably, CFR+SBRL-HAP can further decrease the average $\text{HSIC}_{\text{RFF}}$ to $0.58$, with $37\%$ reduction compared to CFR. Since the major difference between CFR+SBRL-HAP and CFR is the feature decorrelation with hierarchical attention, we can safely conclude that such feature decorrelation can promote the model to identify stable features and acquire more effective associations with potential outcomes, thus enhancing the generalization ability.

\subsubsection{Ablation Studies}

Table~\ref{tab:ablation} reports the effects of each sub-module of our SBRL-HAP by conducting ablation experiments on $\text{Syn}\_16\_16\_16\_2$ dataset. The observations are as follows:
(1) Each component of our SBRL-HAP is indispensable since the absence of any one of them would hinder obtaining balanced and stable representations and damage the performance of HTE estimation on OOD data.
(2) Compared to IR and BR, HAP has the greatest impact on the model's performance on OOD populations of $\text{Syn}\_16\_16\_16\_2$ data.

\subsection{Experiments on Real-world Data}
\subsubsection{Datasets}
We also conduct experiments on two real-world datasets, Twins and IHDP, which are widely used in HTE estimation literature \cite{hassanpourCounterFactualRegressionImportance2019,yaoACEAdaptivelySimilarityPreserved2019,shalitEstimatingIndividualTreatment2017}.

% TWINS
\textbf{Twins}\footnote{ http://www.nber.org/data/linked-birth-infant-death-data-vital-statistics-data.html.}. The Twins dataset originates from twins birth in the USA between 1989 and 1991 \cite{almondCostsLowBirth2005}.
The treatment corresponds to twins' weight, where $t=1$ indicates the heavier twin and $t=0$ indicates the lighter one.
The outcome corresponds to the twins' mortality after one year.
We collect records of same-sex twins weighing less than $2000g$ and without missing features, resulting in a total of 5271 records.
The dataset consists of 43 variables $X=\{X_1,X_2,\ldots,X_{43}\}$, of which $X_C=\{X_1,X_2,\ldots,X_{28}\}$ are derived from the original data related to parents, pregnancy, and birth.
In addition, 10 instrumental variables $X_I=\{X_{29},X_{30},\ldots,X_{38}\}$ and 5 unstable variables $X_V=\{X_{39},X_{40},\ldots,X_{43}\}$ are generated with normal distribution $\mathcal{N}(0,1)$.
To simulate selection bias, treatment is assigned as follows: $t_i|x_i\sim \mathcal{B}(\frac{1}{1+e^{-z}})$, where $z=w^TX_{IC}+\eta$, $w\sim \mathcal{U}(-0.1,0.1)$ and $\eta\sim \mathcal{N}(0,0.1)$.
$\mathcal{B}$ denotes the Bernoulli distribution.
Besides, to create distribution shift, we generate selection probabilities for each sample in the following way: $\Pr=\prod_{X_i\in X_V}|\rho|^{-10*D_i}$, where $D_i=|Y_1-Y_0-\text{sign}(\rho)*X_i|$. Here, we set $\rho=-2.5$.
Based on the sample probabilities, $20\%$ records are sampled as the testing set. Then, the rest data is randomly split into a training/validation set using a 70/30 ratio.
Repeat the above data partitioning for 10 rounds to form the final dataset.

% Table generated by Excel2LaTeX from sheet 'ablation-study'
\begin{table}[!tbp]
  \centering
  \caption{Ablation experiments on the performance of each sub-module. ($\checkmark$ refers to keeping the sub-module.)}
  \begin{threeparttable}
  \begingroup %这一行要加！！！！！！
  \setlength{\tabcolsep}{6pt} % 调整列间距  Default value: 6pt
  \renewcommand{\arraystretch}{0.8} % 调整行间距  Default value: 1
    \begin{tabular}{ccc|cc}
    \toprule
    \multirow{2}[4]{*}{BR ($\mathcal{L}_{\textbf{B}}$)} & \multirow{2}[4]{*}{IR ($\mathcal{L}_{\textbf{I}}$)} & \multirow{2}[4]{*}{HAP ($\mathcal{L}_{\textbf{H}}$)} & \multicolumn{2}{c}{PEHE} \\
\cmidrule{4-5}        &     &     & $\rho=2.5$ & $\rho=-3$ \\
    \midrule
        & $\checkmark$ & $\checkmark$ & 0.457$\pm$0.006 & 0.594$\pm$0.002 \\
    \midrule
    $\checkmark$ &     & $\checkmark$ & 0.502$\pm$0.007 & 0.584$\pm$0.006 \\
    \midrule
    $\checkmark$ & $\checkmark$ &     & \boldmath{}\textbf{0.439$\pm$0.006}\unboldmath{} & 0.662$\pm$0.015 \\
    \midrule
    $\checkmark$ & $\checkmark$ & $\checkmark$ & 0.460$\pm$0.007 & \boldmath{}\textbf{0.591$\pm$0.004}\unboldmath{} \\
    \bottomrule
    \end{tabular}%
  \label{tab:ablation}%
  \endgroup
  \begin{tablenotes}
    \item * $\mathcal{L}_{\textbf{H}}=\mathcal{L}_{\textbf{D}}(\mathbf{Z}^r,\mathbf{w})+\mathcal{L}_{\textbf{D}}(\mathbf{Z}^o,\mathbf{w})$.
  \end{tablenotes}
  \end{threeparttable}
\end{table}%

% IHDP
\textbf{IHDP}\footnote{http://www.fredjo.com.}. 
This is a binary-treatment and continuous-outcome dataset, generated from the Randomized Controlled Trial (RCT) data of the Infant Health and Development Program (IHDP)~\cite{hillBayesianNonparametricModeling2011}. The RCT data of IHDP is collected to evaluate the effect of specialist home visits on the cognitive test scores of premature infants. Hill induced selection bias by removing a biased subset of the treated group, and Shuilte simulated outcomes by setting ``A'' of the NPCI package~\cite{dorieVdorieNpci2023}. This dataset contains 747 units (139 treated, 608 control) with 25 covariates (6 continuous, 19 discrete) related to children and mothers.
To introduce distribution shift, we biasedly sample $10\%$ records as the testing set with specific selection probabilities $\Pr=\prod_{X_i\in X_l}|\rho|^{-10*D_i}$, where $X_l$ are continuous variables, and $D_i=|Y_1-Y_0-\text{sign}(\rho)*X_i|$. The remaining $90\%$ of records are divided randomly into training/validation with a 70/30 proportion.
Different from the unstable variables $X_V \sim \mathcal{N}(0,1)$ in Twins, we choose the subset of original variables in IHDP to introduce distribution shift. This approach aims to create a more complex scenario to verify the effectiveness of our method.

% Table generated by Excel2LaTeX from sheet 'IHDP&Twins'
\begin{table*}[htbp]
  \centering
  \caption{The results (mean$\pm$std) of treatment effect estimation on real-world data. Our methods significantly improve the accuracy of HTE estimation on the testing set, with comparable performances on the training set compared to the baselines.}
  \begingroup %这一行要加！！！！！！
  \setlength{\tabcolsep}{6pt} % 调整列间距  Default value: 6pt
  \renewcommand{\arraystretch}{0.8} % 调整行间距  Default value: 1
    \begin{tabular}{lccc|ccc}
    \toprule
        & \multicolumn{6}{c}{\textbf{Twins}} \\
    \midrule
    Metric & \multicolumn{3}{c|}{\boldmath{}\textbf{PEHE (Mean$\pm$Std)}\unboldmath{}} & \multicolumn{3}{c}{\boldmath{}\textbf{$\bm{\epsilon_{\text{ATE}}}$ (Mean$\pm$Std)}\unboldmath{}} \\
    \midrule
    Dataset & Training & Validation & Testing & Training & Validation & \multicolumn{1}{c}{Testing} \\
    \midrule
    TARNet & 0.313$\pm$0.010 & 0.342$\pm$0.014 & 0.630$\pm$0.012 & \boldmath{}\textbf{0.024$\pm$0.005}\unboldmath{} & \boldmath{}\textbf{0.028$\pm$0.007}\unboldmath{} & 0.355$\pm$0.007 \\
    +SBRL & 0.309$\pm$0.011 & 0.336$\pm$0.014 & 0.621$\pm$0.009 & 0.026$\pm$0.004 & 0.031$\pm$0.006 & 0.348$\pm$0.004 \\
    +SBRL-HAP & \boldmath{}\textbf{0.236$\pm$0.006}\unboldmath{} & \boldmath{}\textbf{0.239$\pm$0.007}\unboldmath{} & \boldmath{}\textbf{0.547$\pm$0.003}\unboldmath{} & 0.057$\pm$0.001 & 0.056$\pm$0.002 & \boldmath{}\textbf{0.321$\pm$0.002}\unboldmath{} \\
    \midrule
    CFR & 0.294$\pm$0.013 & 0.313$\pm$0.018 & 0.613$\pm$0.012 & 0.024$\pm$0.004 & 0.025$\pm$0.005 & 0.352$\pm$0.005 \\
    +SBRL & 0.287$\pm$0.014 & 0.307$\pm$0.018 & 0.611$\pm$0.013 & \boldmath{}\textbf{0.020$\pm$0.005}\unboldmath{} & \boldmath{}\textbf{0.023$\pm$0.006}\unboldmath{} & 0.356$\pm$0.006 \\
    +SBRL-HAP & \boldmath{}\textbf{0.236$\pm$0.005}\unboldmath{} & \boldmath{}\textbf{0.238$\pm$0.007}\unboldmath{} & \boldmath{}\textbf{0.547$\pm$0.003}\unboldmath{} & 0.056$\pm$0.001 & 0.056$\pm$0.002 & \boldmath{}\textbf{0.321$\pm$0.001}\unboldmath{} \\
    \midrule
    DeRCFR & 0.229$\pm$0.002 & 0.229$\pm$0.003 & 0.585$\pm$0.009 & 0.041$\pm$0.013 & 0.040$\pm$0.013 & 0.385$\pm$0.013 \\
    +SBRL & \boldmath{}\textbf{0.229$\pm$0.002}\unboldmath{} & \boldmath{}\textbf{0.229$\pm$0.003}\unboldmath{} & 0.584$\pm$0.009 & \boldmath{}\textbf{0.040$\pm$0.013}\unboldmath{} & \boldmath{}\textbf{0.039$\pm$0.013}\unboldmath{} & 0.384$\pm$0.013 \\
    +SBRL-HAP & 0.236$\pm$0.002 & 0.236$\pm$0.004 & \boldmath{}\textbf{0.552$\pm$0.006}\unboldmath{} & 0.048$\pm$0.010 & 0.047$\pm$0.011 & \boldmath{}\textbf{0.330$\pm$0.011}\unboldmath{} \\
    \midrule
    \midrule
        & \multicolumn{6}{c}{\textbf{IHDP}} \\
    \midrule
    Metric & \multicolumn{3}{c|}{\boldmath{}\textbf{PEHE (Mean$\pm$Std)}\unboldmath{}} & \multicolumn{3}{c}{\boldmath{}\textbf{$\bm{\epsilon_{\text{ATE}}}$ (Mean$\pm$Std)}\unboldmath{}} \\
    \midrule
    Dataset & Training & Validation & Testing & Training & Validation & \multicolumn{1}{c}{Testing} \\
    \midrule
    TARNet & \boldmath{}\textbf{0.620$\pm$0.042}\unboldmath{} & \boldmath{}\textbf{0.677$\pm$0.056}\unboldmath{} & 0.857$\pm$0.098 & 0.200$\pm$0.026 & 0.199$\pm$0.026 & 0.254$\pm$0.037 \\
    +SBRL & 0.622$\pm$0.042 & 0.683$\pm$0.057 & 0.834$\pm$0.093 & 0.184$\pm$0.025 & 0.183$\pm$0.025 & 0.250$\pm$0.037 \\
    +SBRL-HAP & 0.628$\pm$0.041 & 0.696$\pm$0.058 & \boldmath{}\textbf{0.827$\pm$0.089}\unboldmath{} & \boldmath{}\textbf{0.179$\pm$0.023}\unboldmath{} & \boldmath{}\textbf{0.179$\pm$0.023}\unboldmath{} & \boldmath{}\textbf{0.226$\pm$0.032}\unboldmath{} \\
    \midrule
    CFR & 0.628$\pm$0.042 & 0.687$\pm$0.057 & 0.858$\pm$0.099 & 0.197$\pm$0.026 & 0.196$\pm$0.026 & 0.259$\pm$0.038 \\
    +SBRL & \boldmath{}\textbf{0.622$\pm$0.043}\unboldmath{} & \boldmath{}\textbf{0.681$\pm$0.059}\unboldmath{} & 0.848$\pm$0.094 & 0.196$\pm$0.027 & 0.197$\pm$0.027 & 0.251$\pm$0.037 \\
    +SBRL-HAP & 0.623$\pm$0.038 & 0.688$\pm$0.053 & \boldmath{}\textbf{0.820$\pm$0.087}\unboldmath{} & \boldmath{}\textbf{0.185$\pm$0.024}\unboldmath{} & \boldmath{}\textbf{0.184$\pm$0.024}\unboldmath{} & \boldmath{}\textbf{0.220$\pm$0.031}\unboldmath{} \\
    \midrule
    DeRCFR & 0.460$\pm$0.024 & 0.487$\pm$0.029 & 0.607$\pm$0.062 & 0.150$\pm$0.022 & 0.152$\pm$0.022 & 0.183$\pm$0.025 \\
    +SBRL & 0.450$\pm$0.022 & \boldmath{}\textbf{0.476$\pm$0.028}\unboldmath{} & 0.592$\pm$0.062 & \boldmath{}\textbf{0.141$\pm$0.019}\unboldmath{} & \boldmath{}\textbf{0.143$\pm$0.019}\unboldmath{} & 0.181$\pm$0.024 \\
    +SBRL-HAP & \boldmath{}\textbf{0.449$\pm$0.023}\unboldmath{} & 0.478$\pm$0.029 & \boldmath{}\textbf{0.573$\pm$0.057}\unboldmath{} & 0.151$\pm$0.021 & 0.154$\pm$0.021 & \boldmath{}\textbf{0.178$\pm$0.024}\unboldmath{} \\
    \bottomrule
    \end{tabular}%
  \endgroup
  \label{tb:semi-syn}%
\end{table*}%

\subsubsection{Results}
We report the mean and standard deviation (std) of treatment effect over 10 replications on Twins and 100 replications on IHDP datasets in Table~\ref{tb:semi-syn}.
The results show that in comparison with state-of-the-art methods, our SBRL achieves significantly better performance on the testing set, while avoiding model overfitting and maintaining similar performance to the baseline methods on the training set.
Especially on Twins, our proposed SBRL-HAP reduces the error metric PEHE by $13.1\%$, $10.8\%$, and $5.6\%$ for TARNet, CFR, and DeR-CFR, as well as minimizes the ATE bias by $9.6\%$, $8.8\%$, and $14.3\%$.

Compared to synthetic datasets, the performance of our method on real-world datasets is enhanced, but the improvement is not stably significant. According to the characteristics and experiment results of Twins and IHDP datasets, we have the following observations. During the training process, the hierarchical independence measure of Twins dataset consistently remains significantly lower compared to the other datasets employed. Since most parents made similar pregnancy preparations, there is an abundance of similar or identical variables in Twins dataset, resulting in distribution differences that are not highly significant in different environments. Although our algorithm eliminated the OOD issue, the level of OOD is too low to indicate remarkable improvement. Similarly, due to limited distribution shift, the performance of our algorithm on IHDP dataset only improved by $2.3\% \sim  15.1\%$. Furthermore, for IHDP dataset, we introduce a more complex covariate shift than the traditional settings~\cite{kuangStablePredictionModel2020,xuTheoreticalAnalysisIndependencedriven2022}: among the six continuous variables used for biased sampling, some may have causation with the outcome $Y$. Artificially introducing unstable correlation on these potentially stable features would make it difficult for the model to identify real stable representations.
 % The analyses are as follows. On Twins dataset, the hierarchical independence loss is remarkably minimal, approximately one-tenth of that observed in synthetic datasets. It, hence, hinders the effectiveness of our model on generalization. As for IHDP dataset, we introduce a more complex covariate shift than the traditional settings~\cite{kuangStablePredictionModel2020,xuTheoreticalAnalysisIndependencedriven2022}: among the six continuous variables used for biased sampling, some may have causation with the outcome $Y$. Artificially introducing unstable correlation on these potentially-stable features would make it difficult for the model to identify the real stable representation.
% On IHDP, the performance of our method on the test set is also enhanced, but the improvement is not stably significant. We attribute this to the following reason: among the six continuous variables used for biased sampling, some may have causation with the outcome $Y$. Artificially introducing unstable correlation on this basis would make it difficult for the model to identify the true stable representation.

% Table generated by Excel2LaTeX from sheet '最优参'
\begin{table*}[htbp]
  \centering
  \caption{Optimal hyper-parameters of CFR+SBRL-HAP.}
    \begingroup %这一行要加！！！！！！
  \setlength{\tabcolsep}{6pt} % 调整列间距  Default value: 6pt
  \renewcommand{\arraystretch}{0.8} % 调整行间距  Default value: 1
    \begin{threeparttable}
    \label{tb:parm-cfr}
    \begin{tabular}{ccccc}
    \toprule
    \textbf{Hyper-parameters} & \textbf{Twins} & \textbf{IHDP} & \textbf{Syn\_8\_8\_8\_2} & \textbf{Syn\_16\_16\_16\_2} \\
    \midrule
    \midrule
    learning rate & 1e-5 & 1e-3 & 1e-5 & 1e-4 \\
    batch norm & 1   & 0   & 1   & 1 \\
    rep normalization & 1   & 1   & 0   & 0 \\
    \midrule
    $\{d_r,d_y\}$ & \{3,3\} & \{3,3\} & \{3,3\} & \{3,3\} \\
    $\{h_r,h_y\}$ & \{128,64\} & \{256,128\} & \{128,64\} & \{128,64\} \\
    $\{\alpha,\lambda\}$ & \{1e-4,1e-4\} & \{1,1e-4\} & \{5e-2,1e-4\} & \{1e-3,1e-4\} \\
    \midrule
    $\{\gamma_1,\gamma_2,\gamma_3\}$ & \{1,1,1e-1\} & \{1e-1,1e-4,1e-4\} & \{1,1,1e-1\} & \{1,1e-3,1e-3\} \\
    \bottomrule
    \end{tabular}%
  \begin{tablenotes}
    \item * Set $\alpha$ to $0$ to get the optimal hyper-parameters of TARNet+SBRL-HAP.
  \end{tablenotes}
  \end{threeparttable}
  \endgroup
\end{table*}%

% Table generated by Excel2LaTeX from sheet '最优参'
\begin{table*}[htbp]
  \centering
  \caption{Optimal hyper-parameters of DeR-CFR+SBRL-HAP.}
  \begingroup %这一行要加！！！！！！
  \setlength{\tabcolsep}{6pt} % 调整列间距  Default value: 6pt
  \renewcommand{\arraystretch}{0.8} % 调整行间距  Default value: 1
  \begin{threeparttable}
  \label{tb:parm-dercfr}
    \begin{tabular}{ccccc}
    \toprule
    \textbf{Hyper-parameters} & \textbf{Twins} & \textbf{IHDP} & \textbf{Syn\_8\_8\_8\_2} & \textbf{Syn\_16\_16\_16\_2} \\
    \midrule
    \midrule
    learning rate & 1e-1 & 1e-3 & 1e-4 & 5e-4 \\
    batch norm & 1   & 0   & 1   & 1 \\
    rep normalization & 1   & 1   & 0   & 0 \\
    \midrule
    $\{d_r,d_y,d_t\}$ & \{3,3,2\} & \{5,3,1\} & \{2,2,3\} & \{2,2,3\} \\
    $\{h_r,h_y,h_t\}$ & \{256,128,128\} & \{32,256,128\} & \{256,256,256\} & \{256,256,256\} \\
    $\{\alpha,\beta,\gamma,\mu,\lambda\}$ & \{1e-2,5,1e-4,5,5\} & \{10,5,1e-3,50,10\} & \{1,1e-3,5,1,1\} & \{1,1e-3,5,1,1\} \\
    \midrule
    $\{\gamma_1,\gamma_2,\gamma_3\}$ & \{1,1,1e-2\} & \{1,1e-1,1e-2\} & \{1,1e-2,1\} & \{1,1e-2,1e-2\} \\
    \bottomrule
    \end{tabular}%
  \begin{tablenotes}
    \item * Refer to DeR-CFR~\cite{wuLearningDecomposedRepresentations2022} for the meaning of hyper-parameters $\{\alpha,\beta,\gamma,\mu,\lambda\}$.
  \end{tablenotes}
  \end{threeparttable}
  \endgroup
\end{table*}%

\subsection{Hyper-parameter Analysis}
Table ~\ref{tb:parm-cfr} and Table~\ref{tb:parm-dercfr} list all optimal hyper-parameters of our SBRL-HAP used for each dataset. Note that setting $\{\gamma_1,\gamma_2,\gamma_3\}$ to $0$ in Table~\ref{tb:parm-cfr} and Table~\ref{tb:parm-dercfr} denotes the optimal hyper-parameters of our SBRL.
Given that the hyper-parameters $\{\gamma_1,\gamma_2,\gamma_3\}$ determine the hierarchical attention for variable decorrelation, we investigate the impact of each hyper-parameter on the model's performance and stability.
As shown in Fig.~\ref{fig:Exp_Hpyer_param}, we report PEHE on data $\text{Syn}\_16\_16\_16\_2$ with $\rho=2.5$ and $F_1$ scores of factual outcomes with $\rho=-3$ by changing $\{\gamma_1,\gamma_2,\gamma_3\}$ in the scope $\{0, 0.01, 0.1, 1, 10, 100\}$. Since PEHE is higher under $\gamma_1=0$ compared to that under $\gamma_1=100$, and PEHE is lower under $\gamma_2=0$ than that under $\gamma_2=100$, we conclude that it is better to give relatively more attention to the last layer of models and comparatively less attention to the balanced representation layer. Besides, compared to $\gamma_1$ and $\gamma_2$, the impact of $\gamma_3$ on the model's performance and stability is more complex.
This is because $\gamma_3$ controls attention to nearly all hidden layers, so that slight modifications in $\gamma_3$ can result in significant changes in the entire loss. Hyper-parameters analysis assists us in identifying the most suitable hyper-parameters for experiments.

\vspace{-1pt}
\subsection{Training Cost Analysis}
In our method, the network structure and hierarchical-attention independence constraints are the primary contributors to the increased model complexity and training time. To investigate the complexity of all methods, we implement 10 replications on IHDP dataset to study the average training time(s) in a single execution, as shown in Table~\ref{tb:time_cost}.
Table~\ref{tb:time_cost} indicates that our SBRL results in nearly twice the training cost than TARNet and CFR. This is due to the additional training process for sample weights compared to TARNet and CFR. Besides, our SBRL-HAP leads to over a 3-fold increase in training time of TARNet and CFR, and a 1.5-fold increase for DeR-CFR. Such an increase is primarily due to the hierarchical-attention optimization strategy. As the model complexity increases, both accuracy and stability of the model improve. Despite its higher computational time, our proposed method achieves the most stable and accurate treatment effect estimation. Fortunately, the maximum training time in a single execution is less than 180 seconds, which is still acceptable.

% Revision
\begin{figure}[!tbp]
	\centering
	\subfigure[The PEHE error with $\rho=2.5$.]{
		\begin{minipage}{0.475\textwidth}
            \includegraphics[width=\textwidth]{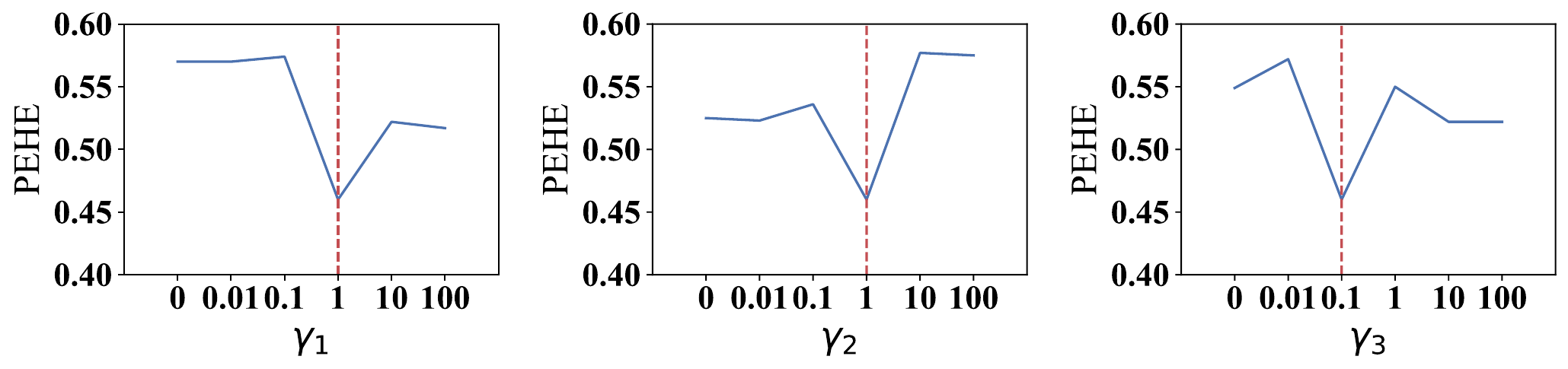}
		\end{minipage}
	}
	\subfigure[The $F_1$ score on factual outcomes with $\rho=-3$.]{
		\begin{minipage}{0.475\textwidth}
			\includegraphics[width=\textwidth]{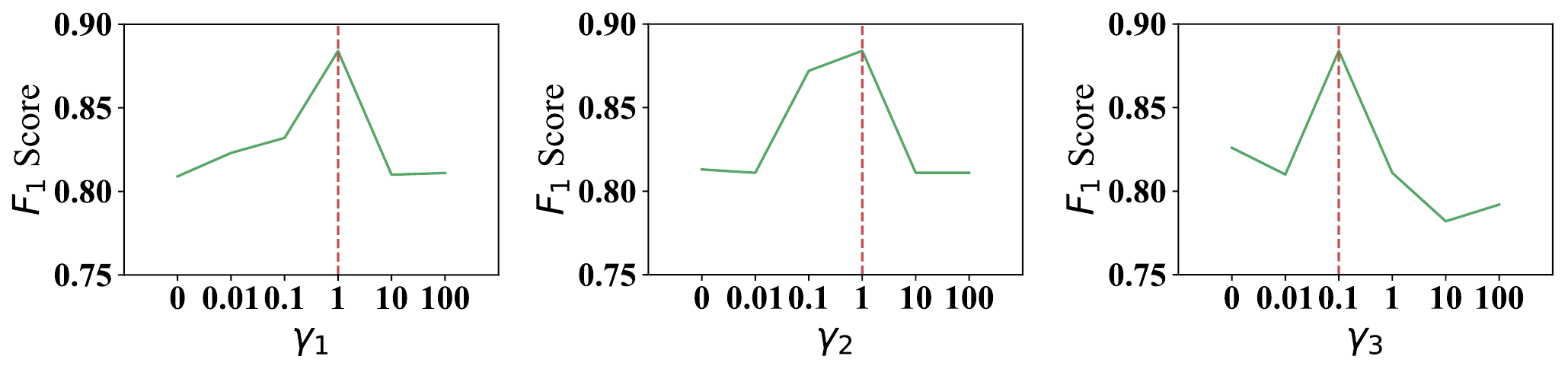}
		\end{minipage}
	}
    \caption{Hyper-parameter sensitivity analysis on $\{\gamma_1,\gamma_2,\gamma_3\}$ within the specified range $\{0, 0.01, 0.1, 1, 10, 100\}$ on $\text{Syn}\_16\_16\_16\_2$ dataset. The reference red line indicates the optimal parameters for the setting.}
    \label{fig:Exp_Hpyer_param}
\setlength{\belowcaptionskip}{-10pt}
\end{figure}

Hardware configuration: CentOS Linux release 7.2 (Final) operating system with the AMD EPYC 7K62 48-Core CPU Processor, 1TB of RAM.
Software configuration: Python~3.6.8 with TensorFlow~1.15.0, NumPy~1.19.5, Scikit-learn~0.24.2.

\section{Conclusion and Future}

In this paper, we first study the problem of the Heterogeneous Treatment Effect across Out-of-distribution Populations. 
Previous causal methods have primarily concentrated on addressing selection bias within in-distribution data. However, in real-world applications, where distribution shifts are common, these methods may face challenges in effectively handling OOD data.
% The previous causal methods mainly focus on selection bias within in-distribution data. However, in real-world applications, where distribution shifts occur frequently, these methods may struggle to perform well across OOD data. 
% In light of this, we propose a Stable Balanced Representation Learning with Hierarchical-Attention Paradigm (SBRL-HAP) to jointly address selection bias and distribution shift by synergistically optimizing a Balancing Regularizer and an Independence Regularizer in a Hierarchical-Attention Paradigm. Empirical results demonstrate the advantages of the SBRL-HAP algorithm compared with state-of-the-art methods.
To achieve more accurate HTE estimation on OOD data, we propose a Stable Balanced Representation Learning with Hierarchical-Attention Paradigm (SBRL-HAP) to jointly address selection bias and distribution shift by synergistically optimizing a Balancing Regularizer and an Independence Regularizer in a Hierarchical-Attention Paradigm.
One limitation is that when combining existing balanced representation methods with SBRL-HAP, the performance on in-distribution data may decrease compared to vanilla methods. Because vanilla methods would rely on the inductive bias from unstable features to improve performance on in-distribution data, which does not generalize well to OOD populations. One potential solution to find a balance between stability and performance is to incorporate a module that measures the OOD level between the target domain and the source domain. Based on the measured OOD level, it would be feasible to use interpolation or spline methods to boost our algorithm with conventional supervised learning, which is left to future work.

% Table generated by Excel2LaTeX from sheet 'time cost'
\begin{table}[!tbp]
  \centering
  \caption{Training time(s) of various methods in a single execution on IHDP dataset.}
  \begingroup %这一行要加！！！！！！
  \setlength{\tabcolsep}{6pt} % 调整列间距  Default value: 6pt
  \renewcommand{\arraystretch}{0.5} % 调整行间距  Default value: 1
    \begin{tabular}{c|ccc}
    \toprule
    \textbf{Method} & \textbf{TARNet} & \textbf{+SBRL} & \textbf{+SBRL-HAP} \\
    \midrule
    \textbf{Time (s)} & 22.4 & 40.6 & 79.7 \\
    \midrule
    \midrule
    \textbf{Method} & \textbf{CFR} & \textbf{+SBRL} & \textbf{+SBRL-HAP} \\
    \midrule
    \textbf{Time (s)} & 25.3 & 40.8 & 80.1 \\
    \midrule
    \midrule
    \textbf{Method} & \textbf{DeR-CFR} & \textbf{+SBRL} & \textbf{+SBRL-HAP} \\
    \midrule
    \textbf{Time (s)} & 96.4 & 112.1 & 140.5 \\
    \bottomrule
    \end{tabular}%
  \label{tb:time_cost}%
  \endgroup
\end{table}%

\section*{Acknowledgment}

This work was supported in part by National Key Research and Development Project of China (Grant No. 2023YFF0905502), Shenzhen Science and Technology Program (Grant No. RCYX20200714114523079 and JCYJ20220818101014030) and National Natural Science Foundation of China (Grant No. 62376243 and U20A20387). We would like to thank Tencent for supporting the research during Yuling Zhang's internship. We also would like to thank Kuaishou for sponsoring the research. Anpeng Wu's research was supported by the China Scholarship Council.

\bibliographystyle{IEEEtran}
\bibliography{SCFR}

\begin{thebibliography}{10}
\providecommand{\url}[1]{#1}
\csname url@samestyle\endcsname
\providecommand{\newblock}{\relax}
\providecommand{\bibinfo}[2]{#2}
\providecommand{\BIBentrySTDinterwordspacing}{\spaceskip=0pt\relax}
\providecommand{\BIBentryALTinterwordstretchfactor}{4}
\providecommand{\BIBentryALTinterwordspacing}{\spaceskip=\fontdimen2\font plus
\BIBentryALTinterwordstretchfactor\fontdimen3\font minus \fontdimen4\font\relax}
\providecommand{\BIBforeignlanguage}[2]{{%
\expandafter\ifx\csname l@#1\endcsname\relax
\typeout{** WARNING: IEEEtran.bst: No hyphenation pattern has been}%
\typeout{** loaded for the language `#1'. Using the pattern for}%
\typeout{** the default language instead.}%
\else
\language=\csname l@#1\endcsname
\fi
#2}}
\providecommand{\BIBdecl}{\relax}
\BIBdecl

\bibitem{chuContinualCausalInference2023}
Z.~Chu, R.~Li, S.~Rathbun, and S.~Li, ``Continual {{Causal Inference}} with {{Incremental Observational Data}},'' Mar. 2023.

\bibitem{aiLBCFLargeScaleBudgetConstrained2022}
M.~Ai, B.~Li, H.~Gong, Q.~Yu, S.~Xue, Y.~Zhang, Y.~Zhang, and P.~Jiang, ``{{LBCF}}: {{A Large-Scale Budget-Constrained Causal Forest Algorithm}},'' in \emph{Proceedings of the {{ACM Web Conference}} 2022}, ser. {{WWW}} '22.\hskip 1em plus 0.5em minus 0.4em\relax {New York, NY, USA}: {Association for Computing Machinery}, Apr. 2022, pp. 2310--2319.

\bibitem{tanUncoveringCausalEffects2022}
Z.~Tan, S.~Zhang, N.~Hong, K.~Kuang, Y.~Yu, J.~Yu, Z.~Zhao, H.~Yang, S.~Pan, J.~Zhou, and F.~Wu, ``Uncovering {{Causal Effects}} of {{Online Short Videos}} on {{Consumer Behaviors}},'' in \emph{Proceedings of the {{Fifteenth ACM International Conference}} on {{Web Search}} and {{Data Mining}}}, ser. {{WSDM}} '22.\hskip 1em plus 0.5em minus 0.4em\relax {New York, NY, USA}: {Association for Computing Machinery}, Feb. 2022, pp. 997--1006.

\bibitem{mengCausalAnalysisUnsatisfying2019}
Y.~Meng, S.~Zhang, Z.~Ye, B.~Wang, Z.~Wang, Y.~Sun, Q.~Liu, S.~Yang, and D.~Pei, ``Causal {{Analysis}} of the {{Unsatisfying Experience}} in {{Realtime Mobile Multiplayer Games}} in the {{Wild}},'' in \emph{2019 {{IEEE International Conference}} on {{Multimedia}} and {{Expo}} ({{ICME}})}, Jul. 2019, pp. 1870--1875.

\bibitem{wagerEstimationInferenceHeterogeneous2018}
S.~Wager and S.~Athey, ``Estimation and {{Inference}} of {{Heterogeneous Treatment Effects}} using {{Random Forests}},'' \emph{Journal of the American Statistical Association}, vol. 113, no. 523, pp. 1228--1242, Jul. 2018.

\bibitem{pearlCausality2009}
J.~Pearl, \emph{Causality}.\hskip 1em plus 0.5em minus 0.4em\relax {Cambridge University Press.}, 2009.

\bibitem{wangSequentialRecommendationUser2023}
Z.~Wang, X.~Chen, R.~Zhou, Q.~Dai, Z.~Dong, and J.-R. Wen, ``Sequential {{Recommendation}} with {{User Causal Behavior Discovery}},'' in \emph{2023 {{IEEE}} 39th {{International Conference}} on {{Data Engineering}} ({{ICDE}})}, Apr. 2023, pp. 28--40.

\bibitem{zhuDCMTDirectEntireSpace2023}
F.~Zhu, M.~Zhong, X.~Yang, L.~Li, L.~Yu, T.~Zhang, J.~Zhou, C.~Chen, F.~Wu, G.~Liu, and Y.~Wang, ``{{DCMT}}: {{A Direct Entire-Space Causal Multi-Task Framework}} for {{Post-Click Conversion Estimation}},'' in \emph{2023 {{IEEE}} 39th {{International Conference}} on {{Data Engineering}} ({{ICDE}})}, Apr. 2023, pp. 3113--3125.

\bibitem{youngmannExplainingConfoundingBias2023}
B.~Youngmann, M.~Cafarella, Y.~Moskovitch, and B.~Salimi, ``On {{Explaining Confounding Bias}},'' in \emph{2023 {{IEEE}} 39th {{International Conference}} on {{Data Engineering}} ({{ICDE}})}, Apr. 2023, pp. 1846--1859.

\bibitem{shenCausalWhatIfHowTo2023}
F.~Shen, K.~Heravi, O.~Gomez, S.~Galhotra, A.~Gilad, S.~Roy, and B.~Salimi, ``Causal {{What-If}} and {{How-To Analysis Using HypeR}},'' in \emph{2023 {{IEEE}} 39th {{International Conference}} on {{Data Engineering}} ({{ICDE}})}, Apr. 2023, pp. 3663--3666.

\bibitem{rosenbaumCentralRolePropensity1983}
P.~R. Rosenbaum and D.~B. Rubin, ``The central role of the propensity score in observational studies for causal effects,'' \emph{Biometrika}, vol.~70, no.~1, pp. 41--55, 1983.

\bibitem{rosenbaumModelBasedDirectAdjustment1987}
P.~R. Rosenbaum, ``Model-{{Based Direct Adjustment}},'' \emph{Journal of the American Statistical Association}, vol.~82, no. 398, pp. 387--394, Jun. 1987.

\bibitem{liMatchingDimensionalityReduction2016}
S.~Li, N.~Vlassis, J.~Kawale, and Y.~Fu, ``Matching via dimensionality reduction for estimation of treatment effects in digital marketing campaigns,'' in \emph{Proceedings of the {{Twenty-Fifth International Joint Conference}} on {{Artificial Intelligence}}}, ser. {{IJCAI}}'16.\hskip 1em plus 0.5em minus 0.4em\relax {New York, New York, USA}: {AAAI Press}, Jul. 2016, pp. 3768--3774.

\bibitem{yaoSurveyCausalInference2021}
L.~Yao, Z.~Chu, S.~Li, Y.~Li, J.~Gao, and A.~Zhang, ``A {{Survey}} on {{Causal Inference}},'' \emph{ACM Transactions on Knowledge Discovery from Data}, vol.~15, no.~5, pp. 1--46, Oct. 2021.

\bibitem{shalitEstimatingIndividualTreatment2017}
U.~Shalit, F.~D. Johansson, and D.~Sontag, ``Estimating individual treatment effect: Generalization bounds and algorithms,'' in \emph{Proceedings of the 34th {{International Conference}} on {{Machine Learning}}}.\hskip 1em plus 0.5em minus 0.4em\relax {PMLR}, Jul. 2017, pp. 3076--3085.

\bibitem{hassanpourCounterFactualRegressionImportance2019}
N.~Hassanpour and R.~Greiner, ``{{CounterFactual Regression}} with {{Importance Sampling Weights}},'' in \emph{Proceedings of the {{Twenty-Eighth International Joint Conference}} on {{Artificial Intelligence}}}.\hskip 1em plus 0.5em minus 0.4em\relax {Macao, China}: {International Joint Conferences on Artificial Intelligence Organization}, Aug. 2019, pp. 5880--5887.

\bibitem{wuLearningDecomposedRepresentations2022}
A.~Wu, J.~Yuan, K.~Kuang, B.~Li, R.~Wu, Q.~Zhu, Y.~T. Zhuang, and F.~Wu, ``Learning {{Decomposed Representations}} for {{Treatment Effect Estimation}},'' \emph{IEEE Transactions on Knowledge and Data Engineering}, pp. 1--1, 2022.

\bibitem{hassanpourLearningDisentangledRepresentations2020}
N.~Hassanpour and R.~Greiner, ``Learning {{Disentangled Representations}} for {{CounterFactual Regression}},'' in \emph{International {{Conference}} on {{Learning Representations}}}, Mar. 2020.

\bibitem{schwabLearningCounterfactualRepresentations2020a}
P.~Schwab, L.~Linhardt, S.~Bauer, J.~M. Buhmann, and W.~Karlen, ``Learning {{Counterfactual Representations}} for {{Estimating Individual Dose-Response Curves}},'' \emph{Proceedings of the AAAI Conference on Artificial Intelligence}, vol.~34, no.~04, pp. 5612--5619, Apr. 2020.

\bibitem{yaoSCISubspaceLearning2021}
L.~Yao, Y.~Li, S.~Li, M.~Huai, J.~Gao, and A.~Zhang, ``{{SCI}}: {{Subspace Learning Based Counterfactual Inference}} for {{Individual Treatment Effect Estimation}},'' in \emph{Proceedings of the 30th {{ACM International Conference}} on {{Information}} \& {{Knowledge Management}}}, ser. {{CIKM}} '21.\hskip 1em plus 0.5em minus 0.4em\relax {New York, NY, USA}: {Association for Computing Machinery}, Oct. 2021, pp. 3583--3587.

\bibitem{yaoACEAdaptivelySimilarityPreserved2019}
L.~Yao, S.~Li, Y.~Li, M.~Huai, J.~Gao, and A.~Zhang, ``{{ACE}}: {{Adaptively Similarity-Preserved Representation Learning}} for {{Individual Treatment Effect Estimation}},'' in \emph{2019 {{IEEE International Conference}} on {{Data Mining}} ({{ICDM}})}, Nov. 2019, pp. 1432--1437.

\bibitem{yaoRepresentationLearningTreatment2018}
------, ``Representation {{Learning}} for {{Treatment Effect Estimation}} from {{Observational Data}},'' in \emph{Advances in {{Neural Information Processing Systems}}}, vol.~31.\hskip 1em plus 0.5em minus 0.4em\relax {Curran Associates, Inc.}, 2018.

\bibitem{zhangDiversePreferenceAugmentation2022}
Y.~Zhang, C.~Li, I.~W. Tsang, H.~Xu, L.~Duan, H.~Yin, W.~Li, and J.~Shao, ``Diverse {{Preference Augmentation}} with {{Multiple Domains}} for {{Cold-start Recommendations}},'' in \emph{2022 {{IEEE}} 38th {{International Conference}} on {{Data Engineering}} ({{ICDE}})}, May 2022, pp. 2942--2955.

\bibitem{caoCrossDomainRecommendationColdStart2022}
J.~Cao, J.~Sheng, X.~Cong, T.~Liu, and B.~Wang, ``Cross-{{Domain Recommendation}} to {{Cold-Start Users}} via {{Variational Information Bottleneck}},'' in \emph{2022 {{IEEE}} 38th {{International Conference}} on {{Data Engineering}} ({{ICDE}})}, May 2022, pp. 2209--2223.

\bibitem{zhouActiveGradualDomain2022}
S.~Zhou, L.~Wang, S.~Zhang, Z.~Wang, and W.~Zhu, ``Active {{Gradual Domain Adaptation}}: {{Dataset}} and {{Approach}},'' \emph{IEEE Transactions on Multimedia}, vol.~24, pp. 1210--1220, 2022.

\bibitem{zhouOnlineContinualAdaptation2022}
S.~Zhou, H.~Zhao, S.~Zhang, L.~Wang, H.~Chang, Z.~Wang, and W.~Zhu, ``Online {{Continual Adaptation}} with {{Active Self-Training}},'' in \emph{Proceedings of {{The}} 25th {{International Conference}} on {{Artificial Intelligence}} and {{Statistics}}}.\hskip 1em plus 0.5em minus 0.4em\relax {PMLR}, May 2022, pp. 8852--8883.

\bibitem{fangGIFDGenerativeGradient2023}
H.~Fang, B.~Chen, X.~Wang, Z.~Wang, and S.-T. Xia, ``{{GIFD}}: {{A Generative Gradient Inversion Method}} with {{Feature Domain Optimization}},'' in \emph{Proceedings of the {{IEEE}}/{{CVF International Conference}} on {{Computer Vision}}}, 2023, pp. 4967--4976.

\bibitem{wuIterativeRefinementMultiSource2023}
H.~Wu, Y.~Yan, G.~Lin, M.~Yang, M.~K. Ng, and Q.~Wu, ``Iterative {{Refinement}} for {{Multi-Source Visual Domain Adaptation}} ({{Extended}} abstract),'' in \emph{2023 {{IEEE}} 39th {{International Conference}} on {{Data Engineering}} ({{ICDE}})}, Apr. 2023, pp. 3829--3830.

\bibitem{yanTransferableFeatureSelection2023}
Y.~Yan, H.~Wu, Y.~Ye, C.~Bi, M.~Lu, D.~Liu, Q.~Wu, and M.~K. Ng, ``Transferable {{Feature Selection}} for {{Unsupervised Domain Adaptation}} : {{Extended Abstract}},'' in \emph{2023 {{IEEE}} 39th {{International Conference}} on {{Data Engineering}} ({{ICDE}})}, Apr. 2023, pp. 3855--3856.

\bibitem{chenUnsupervisedIntraDomainAdaptation2023}
C.~Chen, J.~Xiao, J.~Liu, J.~Zhang, J.~Jia, and N.~Hu, ``Unsupervised {{Intra-Domain Adaptation}} for {{Recommendation}} via {{Uncertainty Minimization}},'' in \emph{2023 {{IEEE}} 39th {{International Conference}} on {{Data Engineering Workshops}} ({{ICDEW}})}, Apr. 2023, pp. 79--86.

\bibitem{chenBAGNNLearningBiasAware2022}
Z.~Chen, T.~Xiao, and K.~Kuang, ``{{BA-GNN}}: {{On Learning Bias-Aware Graph Neural Network}},'' in \emph{2022 {{IEEE}} 38th {{International Conference}} on {{Data Engineering}} ({{ICDE}})}, May 2022, pp. 3012--3024.

\bibitem{yuanLabelEfficientDomainGeneralization2022}
J.~Yuan, X.~Ma, D.~Chen, K.~Kuang, F.~Wu, and L.~Lin, ``Label-{{Efficient Domain Generalization}} via {{Collaborative Exploration}} and {{Generalization}},'' in \emph{Proceedings of the 30th {{ACM International Conference}} on {{Multimedia}}}, ser. {{MM}} '22.\hskip 1em plus 0.5em minus 0.4em\relax {New York, NY, USA}: {Association for Computing Machinery}, Oct. 2022, pp. 2361--2370.

\bibitem{sugiyamaCovariateShiftAdaptation2007}
M.~Sugiyama, M.~Krauledat, and K.-R. M{\"u}ller, ``Covariate {{Shift Adaptation}} by {{Importance Weighted Cross Validation}},'' \emph{The Journal of Machine Learning Research}, vol.~8, pp. 985--1005, Dec. 2007.

\bibitem{shimodairaImprovingPredictiveInference2000}
H.~Shimodaira, ``Improving predictive inference under covariate shift by weighting the log-likelihood function,'' \emph{Journal of Statistical Planning and Inference}, vol.~90, no.~2, pp. 227--244, Oct. 2000.

\bibitem{cuiStableLearningEstablishes2022}
P.~Cui and S.~Athey, ``Stable learning establishes some common ground between causal inference and machine learning,'' \emph{Nature Machine Intelligence}, vol.~4, no.~2, pp. 110--115, Feb. 2022.

\bibitem{wangLearningRobustRepresentations2019a}
H.~Wang, Z.~He, Z.~C. Lipton, and E.~P. Xing, ``Learning {{Robust Representations}} by {{Projecting Superficial Statistics Out}},'' Mar. 2019.

\bibitem{muandetDomainGeneralizationInvariant2013a}
K.~Muandet, D.~Balduzzi, and B.~Sch{\"o}lkopf, ``Domain {{Generalization}} via {{Invariant Feature Representation}},'' in \emph{Proceedings of the 30th {{International Conference}} on {{Machine Learning}}}.\hskip 1em plus 0.5em minus 0.4em\relax {PMLR}, Feb. 2013, pp. 10--18.

\bibitem{fanGeneralizingGraphNeural2021a}
S.~Fan, X.~Wang, C.~Shi, P.~Cui, and B.~Wang, ``Generalizing {{Graph Neural Networks}} on {{Out-Of-Distribution Graphs}},'' Nov. 2021.

\bibitem{zhangDeepStableLearning2021}
X.~Zhang, P.~Cui, R.~Xu, L.~Zhou, Y.~He, and Z.~Shen, ``Deep {{Stable Learning}} for {{Out-of-Distribution Generalization}},'' in \emph{Proceedings of the {{IEEE}}/{{CVF Conference}} on {{Computer Vision}} and {{Pattern Recognition}}}, 2021, pp. 5372--5382.

\bibitem{lakeBuildingMachinesThat2017}
B.~M. Lake, T.~D. Ullman, J.~B. Tenenbaum, and S.~J. Gershman, ``Building machines that learn and think like people,'' \emph{Behavioral and Brain Sciences}, vol.~40, p. e253, Jan. 2017.

\bibitem{wuLearningInstrumentalVariable2023}
A.~Wu, K.~Kuang, R.~Xiong, M.~Zhu, Y.~Liu, B.~Li, F.~Liu, Z.~Wang, and F.~Wu, ``Learning {{Instrumental Variable}} from {{Data Fusion}} for {{Treatment Effect Estimation}},'' \emph{Proceedings of the AAAI Conference on Artificial Intelligence}, vol.~37, no.~9, pp. 10\,324--10\,332, Jun. 2023.

\bibitem{johanssonLearningRepresentationsCounterfactual2016}
F.~Johansson, U.~Shalit, and D.~Sontag, ``Learning {{Representations}} for {{Counterfactual Inference}},'' in \emph{Proceedings of {{The}} 33rd {{International Conference}} on {{Machine Learning}}}.\hskip 1em plus 0.5em minus 0.4em\relax {PMLR}, Jun. 2016, pp. 3020--3029.

\bibitem{johanssonLearningWeightedRepresentations2018}
F.~D. Johansson, N.~Kallus, U.~Shalit, and D.~Sontag, ``Learning {{Weighted Representations}} for {{Generalization Across Designs}},'' Feb. 2018.

\bibitem{changInformativeSubspaceLearning2017}
Y.~Chang and J.~Dy, ``Informative {{Subspace Learning}} for {{Counterfactual Inference}},'' \emph{Proceedings of the AAAI Conference on Artificial Intelligence}, vol.~31, no.~1, Feb. 2017.

\bibitem{kuangEstimatingTreatmentEffect2017}
K.~Kuang, P.~Cui, B.~Li, M.~Jiang, and S.~Yang, ``Estimating {{Treatment Effect}} in the {{Wild}} via {{Differentiated Confounder Balancing}},'' in \emph{Proceedings of the 23rd {{ACM SIGKDD International Conference}} on {{Knowledge Discovery}} and {{Data Mining}}}, ser. {{KDD}} '17.\hskip 1em plus 0.5em minus 0.4em\relax {New York, NY, USA}: {Association for Computing Machinery}, Aug. 2017, pp. 265--274.

\bibitem{atheyApproximateResidualBalancing2018}
S.~Athey, G.~W. Imbens, and S.~Wager, ``Approximate {{Residual Balancing}}: {{De-Biased Inference}} of {{Average Treatment Effects}} in {{High Dimensions}},'' Jan. 2018.

\bibitem{hainmuellerEntropyBalancingCausal2012}
J.~Hainmueller, ``Entropy {{Balancing}} for {{Causal Effects}}: {{A Multivariate Reweighting Method}} to {{Produce Balanced Samples}} in {{Observational Studies}},'' \emph{Political Analysis}, vol.~20, no.~1, pp. 25--46, 2012/ed.

\bibitem{shenCausallyRegularizedLearning2018}
Z.~Shen, P.~Cui, K.~Kuang, B.~Li, and P.~Chen, ``Causally {{Regularized Learning}} with {{Agnostic Data Selection Bias}},'' in \emph{Proceedings of the 26th {{ACM}} International Conference on {{Multimedia}}}, ser. {{MM}} '18.\hskip 1em plus 0.5em minus 0.4em\relax {New York, NY, USA}: {Association for Computing Machinery}, Oct. 2018, pp. 411--419.

\bibitem{kuangStablePredictionModel2020}
K.~Kuang, R.~Xiong, P.~Cui, S.~Athey, and B.~Li, ``Stable {{Prediction}} with {{Model Misspecification}} and {{Agnostic Distribution Shift}},'' \emph{Proceedings of the AAAI Conference on Artificial Intelligence}, vol.~34, no.~04, pp. 4485--4492, Apr. 2020.

\bibitem{shenStableLearningSample2020}
Z.~Shen, P.~Cui, T.~Zhang, and K.~Kunag, ``Stable {{Learning}} via {{Sample Reweighting}},'' \emph{Proceedings of the AAAI Conference on Artificial Intelligence}, vol.~34, no.~04, pp. 5692--5699, Apr. 2020.

\bibitem{guidowimbensCausalInferenceStatistics2015}
{Guido W Imbens} and {Donald B Rubin}, \emph{Causal Inference in Statistics, Social, and Biomedical Sciences}.\hskip 1em plus 0.5em minus 0.4em\relax {Cambridge University Press}, 2015.

\bibitem{liuHeterogeneousRiskMinimization2021}
J.~Liu, Z.~Hu, P.~Cui, B.~Li, and Z.~Shen, ``Heterogeneous {{Risk Minimization}},'' in \emph{Proceedings of the 38th {{International Conference}} on {{Machine Learning}}}.\hskip 1em plus 0.5em minus 0.4em\relax {PMLR}, Jul. 2021, pp. 6804--6814.

\bibitem{xuTheoreticalAnalysisIndependencedriven2022}
R.~Xu, X.~Zhang, Z.~Shen, T.~Zhang, and P.~Cui, ``A {{Theoretical Analysis}} on {{Independence-driven Importance Weighting}} for {{Covariate-shift Generalization}},'' Jul. 2022.

\bibitem{sriperumbudurIntegralProbabilityMetrics2009}
B.~K. Sriperumbudur, K.~Fukumizu, A.~Gretton, B.~Sch{\"o}lkopf, and G.~R.~G. Lanckriet, ``On integral probability metrics, {\textbackslash}phi-divergences and binary classification,'' Oct. 2009.

\bibitem{mullerIntegralProbabilityMetrics1997}
A.~M{\"u}ller, ``Integral {{Probability Metrics}} and {{Their Generating Classes}} of {{Functions}},'' \emph{Advances in Applied Probability}, vol.~29, no.~2, pp. 429--443, Jun. 1997.

\bibitem{grettonKernelStatisticalTest2007}
A.~Gretton, K.~Fukumizu, C.~Teo, L.~Song, B.~Sch{\"o}lkopf, and A.~Smola, ``A {{Kernel Statistical Test}} of {{Independence}},'' in \emph{Advances in {{Neural Information Processing Systems}}}, vol.~20.\hskip 1em plus 0.5em minus 0.4em\relax {Curran Associates, Inc.}, 2007.

\bibitem{stroblApproximateKernelBasedConditional2019}
E.~V. Strobl, K.~Zhang, and S.~Visweswaran, ``Approximate {{Kernel-Based Conditional Independence Tests}} for {{Fast Non-Parametric Causal Discovery}},'' \emph{Journal of Causal Inference}, vol.~7, no.~1, Mar. 2019.

\bibitem{hillBayesianNonparametricModeling2011}
J.~L. Hill, ``Bayesian {{Nonparametric Modeling}} for {{Causal Inference}},'' \emph{Journal of Computational and Graphical Statistics}, vol.~20, no.~1, pp. 217--240, Jan. 2011.

\bibitem{kunschJackknifeBootstrapGeneral1989}
H.~R. Kunsch, ``The {{Jackknife}} and the {{Bootstrap}} for {{General Stationary Observations}},'' \emph{The Annals of Statistics}, vol.~17, no.~3, pp. 1217--1241, 1989.

\bibitem{duchiAdaptiveSubgradientMethods2011}
J.~Duchi, E.~Hazan, and Y.~Singer, ``Adaptive {{Subgradient Methods}} for {{Online Learning}} and {{Stochastic Optimization}},'' \emph{The Journal of Machine Learning Research}, vol.~12, no. null, pp. 2121--2159, Jul. 2011.

\bibitem{bergstraRandomSearchHyperparameter2012}
J.~Bergstra and Y.~Bengio, ``Random search for hyper-parameter optimization,'' \emph{The Journal of Machine Learning Research}, vol.~13, no. null, pp. 281--305, Feb. 2012.

\bibitem{zhangFreeLunchDomain2023}
Y.~Zhang, X.~Wang, J.~Liang, Z.~Zhang, L.~Wang, R.~Jin, and T.~Tan, ``Free {{Lunch}} for {{Domain Adversarial Training}}: {{Environment Label Smoothing}},'' Jan. 2023.

\bibitem{tanProvablyInvariantLearning2023}
X.~Tan, L.~Yong, S.~Zhu, C.~Qu, X.~Qiu, X.~Yinghui, P.~Cui, and Y.~Qi, ``Provably {{Invariant Learning}} without {{Domain Information}},'' in \emph{Proceedings of the 40th {{International Conference}} on {{Machine Learning}}}.\hskip 1em plus 0.5em minus 0.4em\relax {PMLR}, Jul. 2023, pp. 33\,563--33\,580.

\bibitem{kruegerOutofDistributionGeneralizationRisk2021a}
D.~Krueger, E.~Caballero, J.-H. Jacobsen, A.~Zhang, J.~Binas, D.~Zhang, R.~L. Priol, and A.~Courville, ``Out-of-{{Distribution Generalization}} via {{Risk Extrapolation}} ({{REx}}),'' in \emph{Proceedings of the 38th {{International Conference}} on {{Machine Learning}}}.\hskip 1em plus 0.5em minus 0.4em\relax {PMLR}, Jul. 2021, pp. 5815--5826.

\bibitem{zhangDomainSpecificRiskMinimization2023}
Y.-F. Zhang, J.~Wang, J.~Liang, Z.~Zhang, B.~Yu, L.~Wang, D.~Tao, and X.~Xie, ``Domain-{{Specific Risk Minimization}} for {{Domain Generalization}},'' in \emph{Proceedings of the 29th {{ACM SIGKDD Conference}} on {{Knowledge Discovery}} and {{Data Mining}}}, ser. {{KDD}} '23.\hskip 1em plus 0.5em minus 0.4em\relax {New York, NY, USA}: {Association for Computing Machinery}, Aug. 2023, pp. 3409--3421.

\bibitem{zhouSparseInvariantRisk2022}
X.~Zhou, Y.~Lin, W.~Zhang, and T.~Zhang, ``Sparse {{Invariant Risk Minimization}},'' in \emph{Proceedings of the 39th {{International Conference}} on {{Machine Learning}}}.\hskip 1em plus 0.5em minus 0.4em\relax {PMLR}, Jun. 2022, pp. 27\,222--27\,244.

\bibitem{zhangMAPBalancedGeneralization2023}
M.~Zhang, J.~Yuan, Y.~He, W.~Li, Z.~Chen, and K.~Kuang, ``{{MAP}}: {{Towards Balanced Generalization}} of {{IID}} and {{OOD}} through {{Model-Agnostic Adapters}},'' in \emph{Proceedings of the {{IEEE}}/{{CVF International Conference}} on {{Computer Vision}}}, 2023, pp. 11\,921--11\,931.

\bibitem{almondCostsLowBirth2005}
D.~Almond, K.~Y. Chay, and D.~S. Lee, ``The {{Costs}} of {{Low Birth Weight}}*,'' \emph{The Quarterly Journal of Economics}, vol. 120, no.~3, pp. 1031--1083, Aug. 2005.

\bibitem{dorieVdorieNpci2023}
V.~Dorie, ``Vdorie/npci,'' May 2023.

\end{thebibliography}
\vspace{12pt}

% \title{Stable Heterogeneous Treatment Effect Estimation across Out-of-Distribution Populations}
% \maketitle
           
\end{document}